\documentclass[pdflatex,sn-mathphys-num]{sn-jnl}

\usepackage{graphicx}%
\usepackage{multirow}%
\usepackage{amsmath,amssymb,amsfonts}%
\usepackage{amsthm}%
\usepackage{mathrsfs}%
\usepackage[title]{appendix}%
\usepackage{xcolor}%
\usepackage{textcomp}%
\usepackage{manyfoot}%
\usepackage{booktabs}%
\usepackage{algorithm}%
\usepackage{algorithmicx}%
\usepackage{algpseudocode}%
\usepackage{listings}%

\usepackage{bm}
\usepackage{xcolor}
\usepackage{hyperref}
\usepackage{subcaption}
\usepackage{placeins}
\usepackage{dblfloatfix}

\theoremstyle{thmstyleone}%
\theoremstyle{thmstyletwo}%

\theoremstyle{thmstylethree}%

\def\BibTeX{{\rm B\kern-.05em{\sc i\kern-.025em b}\kern-.08em
    T\kern-.1667em\lower.7ex\hbox{E}\kern-.125emX}}
\AtBeginDocument{\definecolor{tmlcncolor}{cmyk}{0.93,0.59,0.15,0.02}\definecolor{NavyBlue}{RGB}{0,86,125}}

\begin{document}

\title[Article Title]{MoSAIK: Motion Search with Artificial Intelligence in K-space}


\author*[1,2]{\fnm{Oscar} \sur{Dabrowski}}
\email{oscar.dabrowski@unige.ch}

\author[3,4]{\fnm{Abijith Jagannath} \sur{Kamath}}

\author[1,2]{\fnm{Frédéric} \sur{Grouiller}}

\author[1,2]{\fnm{Sébastien} \sur{Courvoisier}}

\affil*[1]{%
  \orgname{CIBM Center for Biomedical Imaging},
  \orgaddress{\city{Geneva}, \country{Switzerland}}
}

\affil[2]{%
  \orgdiv{Department of Radiology and Medical Informatics},
  \orgname{Faculty of Medicine, University of Geneva},
  \orgaddress{\city{Geneva}, \country{Switzerland}}
}

\affil[3]{%
  \orgdiv{Biomedical Imaging Group},
  \orgname{École polytechnique fédérale de Lausanne (EPFL)},
  \orgaddress{\city{Lausanne}, \country{Switzerland}}
}

\affil[4]{%
  \orgdiv{Signal Processing EPFL},
  \orgname{CIBM Center for Biomedical Imaging},
  \orgaddress{\city{Lausanne}, \country{Switzerland}}
}


\abstract{

\textbf{Object.}
Motion artefacts remain a major limitation of routine two-dimensional T2-weighted turbo spin-echo (TSE) brain MRI, reducing diagnostic confidence and often necessitating repeated acquisitions. MoSAIK (Motion Search with AI in K-space) retrospectively estimates in-plane rigid-body motion directly from standard multi-coil TSE k-space. Without a motion-free reference, external navigators, or additional acquisitions, MoSAIK infers motion from k-space self-consistency violations and incorporates predicted parameters into model-based, motion-aware reconstruction rather than synthesizing corrected images.

\textbf{Materials and Methods.}
Selected complex coil data were represented as log-magnitude and phase channels. A decoupled convolutional embedding and Transformer encoder estimated phase-encoding-line-indexed rotation and two-dimensional translations. Training used motion-free clinical acquisitions with simulated motion; four held-out subjects were evaluated. Component-wise errors assessed estimation; SSIM and PSNR in silico and entropy focus criterion (EFC) in vivo assessed reconstruction.

\textbf{Results.}
Motion estimation errors were low for both rotation (MSE $<$ 0.2 deg$^2$) and translation (MSE $<$ 0.2 pixels$^2$), with Pearson correlations of approximately 0.96 and 0.94, respectively. In every in-vivo acquisition, more than 75\% of slices showed information gain relative to uncorrected reconstructions. MoSAIK reduced visible artefacts while preserving anatomical detail.


\textbf{Discussion.}
Transformer-based k-space motion-trajectory estimation enables fast retrospective correction of standard clinical TSE brain MRI.

}




\maketitle

\section{Introduction}\label{sec1}

Magnetic resonance imaging (MRI) is strongly established as the most sensitive modality for imaging soft tissue in the brain, but its long scan times make it vulnerable to motion artefacts. This can result in image quality that is either degraded or entirely non-diagnostic, resulting in patient discomfort from repeated scans or, in more severe cases, full exam rescheduling, ultimately leading to increased healthcare costs \cite{andre2015toward, slipsager2020quantifying}.

A wide variety of MRI pulse sequences exist. We focus on brain imaging using the 2D T2-weighted (T2w) multi-slice Turbo/Fast Spin Echo (TSE/FSE) sequence, which is highly sensitive to motion and lacks a clinically viable retrospective correction method, despite being a ''clinical workhorse'' \cite{hennig1986rare}. 

Existing motion correction techniques can be broadly classified as \textit{prospective} or \textit{retrospective}. The former requires updating the MRI pulse sequence in real time based on motion parameters detected by a motion-tracking system \cite{maclaren2013prospective}. We focus on retrospective motion correction, which is independent of any MRI manufacturer and does not rely on any additional hardware or calibration procedures.

For completeness, we briefly mention that motion-tracking methods using external hardware such as cameras or electromagnetic trackers have been explored, but they increase setup complexity (hence require more time), and may be incompatible with setups like 7T scanners with full-coverage head coils \cite{tremblay2005retrospective, afacan2020retrospective, bhuiyan2023feasibility,bhuiyan2024tracking}. As a result, they are rarely, if ever, used in clinical practice.

Another approach uses navigator echoes, which require modifying the pulse sequence by adding extra gradient and RF pulses at each TR to repeatedly sample the same k-space region \cite{ehman1989adaptive, johnson2016retrospective}. Navigators enable retrospective or prospective motion correction, but they increase scan time. This approach is typically not used with TSE sequences in the clinical workflow. 

Motion-robust sequences for T2w imaging exist, such as PROPELLER, which uses radial k-space sampling with so-called “blades” or groups of overlapping radial lines \cite{pipe1999motion}. This approach oversamples the k-space center, reducing sensitivity to motion through averaging and enabling detection of rotations and translations via cross-correlation between blades. However, image quality is generally inferior to Cartesian T2w TSE, and radiologists report that PROPELLER is typically reserved for uncooperative or high-motion pediatric patients. Moreover, PROPELLER requires longer scan times than Cartesian TSE to achieve equivalent k-space coverage due to inefficient peripheral sampling.

Non–machine learning approaches, such as the ''autofocusing'' algorithms introduced by Atkinson \textit{et al.} in the late 1990s, estimate motion by iteratively minimizing an image-based quality metric, typically Shannon entropy \cite{atkinson1997automatic, atkinson1999automatic}. The image is reconstructed using Fourier theorems, and the motion parameter search is guided entirely by the entropy metric in image space. In practice, these methods are often unstable and computationally expensive due to the iterative optimization process. More recent work on autofocusing was proposed in \cite{loktyushin2013blind, loktyushin2015blind}.

Among image-based approaches, several deep learning (DL) methods have been proposed. Image-to-image encoder-decoder convolutional neural networks (CNNs)/U-net-based approaches, such as \cite{pawar2022suppressing, kang2024motion} or generative adversarial-based methods, such as \cite{johnson2019conditional, kustner2019retrospective, chen2021retrospective, armanious2020medgan}, while sometimes producing qualitatively and quantitatively convincing results, risk introducing hallucinated or oversmoothed features, which is particularly problematic in the medical domain, where such artefacts can lead to incorrect diagnosis \cite{bhadra2021hallucinations,gottschling2025troublesome}. Image-based approaches such as NAMER combine CNNs with iterative model-based algorithms to estimate motion parameters and reconstruct motion-mitigated images \cite{haskell2019network}. These methods offer robustness against hallucinations through explicit data consistency constraints, but their high computational cost makes them impractical for routine clinical use, where radiologists rely on having images available within minutes after acquisition. Recent work has begun to explore Vision Transformer (ViT) architectures, which have demonstrated strong performance in computer vision tasks. Zhang \textit{et al.} applied ViTs to motion correction in brain MRI \cite{zhang2024reducing}, using a self-supervised pretraining strategy that involves masking parts of the input images to improve subsequent supervised learning. While the approach is innovative, the network ultimately predicts motion-corrected images directly, which may carry the risk of introducing hallucinations. Recently, a novel deep learning-based motion correction approach targeting TSE sequences was proposed in \cite{lee2024unsupervised}. It does not require any training datasets and relies on ''deep image priors'', which are implicit biases inherent to convolutional architectures \cite{dmitry2020deep}. The authors themselves note that, despite impressive results, the method is computationally intensive, requiring thousands of iterative optimization steps per image. Indeed, since their deep neural networks are not pre-trained, optimization has to be performed at inference.

An alternative to image-domain methods is to operate directly in the raw acquisition domain, i.e., k-space. This paradigm shift offers the advantage of precise motion localization in k-space, due to the inherently sequential nature of MRI data acquisition, whereas in image space, motion effects are spatially distributed. We have previously shown with Search in segmented motion input in k-space (SISMIK) that 2D rigid-body motion parameters can be estimated directly from k-space with a deep neural network, in a relative manner (i.e., motion between phase encoding line $n$ and line $n+1$) without requiring a motion-free reference \cite{dabrowski2024sismik}. Another advantage of working in k-space is increased robustness to variations in image contrast and, to some extent, anatomical content. The method proposed by Hossbach \textit{et al.} operates in k-space but is limited to the central 64 $\times$ 64 region and relies on a motion-free scout acquisition as a reference, which may be impractical in clinical settings\cite{hossbach2023deep}. As in SISMIK, their image reconstruction relies on a model-based approach leveraging Fourier shift and rotation theorems, thereby avoiding any risk of hallucinations.

Eichhorn \textit{et al.} proposed a CNN to detect motion-corrupted k-space lines \cite{eichhorn2023physics}. The model takes corrupted k-space data as input and outputs a binary mask identifying the affected lines. While the approach is promising, it has only been evaluated in silico and remains to be validated in vivo. It also remains to be shown whether it can be extended to estimate motion parameters, rather than solely identifying corrupted lines. 

Perhaps the most recent and advanced k-space-based method for motion correction of 2D TSE sequences is presented in \cite{singh2024data}. Based on a 2D in-plane rigid-body head motion model, they train a hypernetwork, that is a network that generates the weights of a reconstruction network which is conditioned on a ''candidate'' motion vector \textit{m}. At inference, they optimize \textit{m} such that the hypernetwork produces suitable weights for the motion corrected reconstruction of a given 2D slice. They include a data consistency term to mitigate the risk of hallucinations. A major drawback of this approach is that it does not directly predict motion parameters but instead relies on an optimization-based search at inference, resulting in a high computational cost of about 5 minutes per 2D slice.

Finally, ''hybrid'' approaches leveraging both image and k-space domain data have been proposed \cite{cui2023motion, safari2025physics}. PI-MoCoNet was proposed in 2025, consisting of two jointly trained U-net-based networks: one to detect corrupted k-space lines and another to correct the image \cite{safari2025physics}. Learning was guided by three losses: reconstruction, perceptual and k-space data consistency loss. However, this method does not estimate motion parameters explicitly and is trained on simulated artefacts applied to already reconstructed DICOM images rather than raw k-space data, which limits the validity of the data consistency term and raises concerns about generalization and potential hallucinations.

Inspired by recent progress in the field of natural language processing as well as computer vision, allowed by the fully attention-based Transformer neural network architecture \cite{vaswani2017attention, DosovitskiyB0WZ21, MaskedAutoencoders2021}, we explore its application to motion parameter estimation in k-space. We propose a deep learning-based approach leveraging Transformer models to retrospectively extract 2D in-plane rigid-body head motion trajectories from raw multi-coil k-space data of TSE acquisitions. To avoid any risk of hallucination, we retain the approach used in \cite{dabrowski2024sismik}, where DL is only used to estimate motion parameters and a model-based approach leveraging the Fourier shift and rotation theorems performs image reconstruction from the estimated motion trajectory. Moreover, we perform ablation studies to demonstrate the contribution of different components of our model, in particular the importance of a custom physics-based convolutional embedding sub-network as well as the role of attention with respect to a fully convolutional neural network (FCNN). To the best of our knowledge, it is the first time that the Transformer architecture is explored and adapted to work directly in k-space to tackle the problem of motion correction in MRI.

The following sections are organized as follows: Section II presents the problem formulation and describes the methodology, including dataset acquisition, motion simulation and artificial neural network training and experiments performed; Section III presents motion estimation and correction results obtained in-silico and in-vivo as well as ablation tests; Section IV consists of a discussion on the results, limitations of the approach and future work and finally Section V concludes this work.

\section{Methods}

\subsection{Overview of the MoSAIK pipeline}


This section describes the MoSAIK pipeline used to estimate and correct motion in two-dimensional TSE MRI. We first provide an overview of the pipeline and formulate the problem by introducing the forward model for TSE acquisitions in the presence of in-plane rigid-body motion. We then describe the input representation and normalization procedure, followed by the MoSAIK deep learning model for motion parameter estimation, its architecture, training strategy, inference procedure, and motion trajectory post-processing. Next, we detail the theory and practical implementation of the model-based reconstruction algorithm used to recover motion-corrected images. Finally, we describe the datasets, experimental setup, and evaluation procedures used to assess motion estimation and motion correction performance.

\begin{figure*}
\centerline{\includegraphics[width=5.25in]{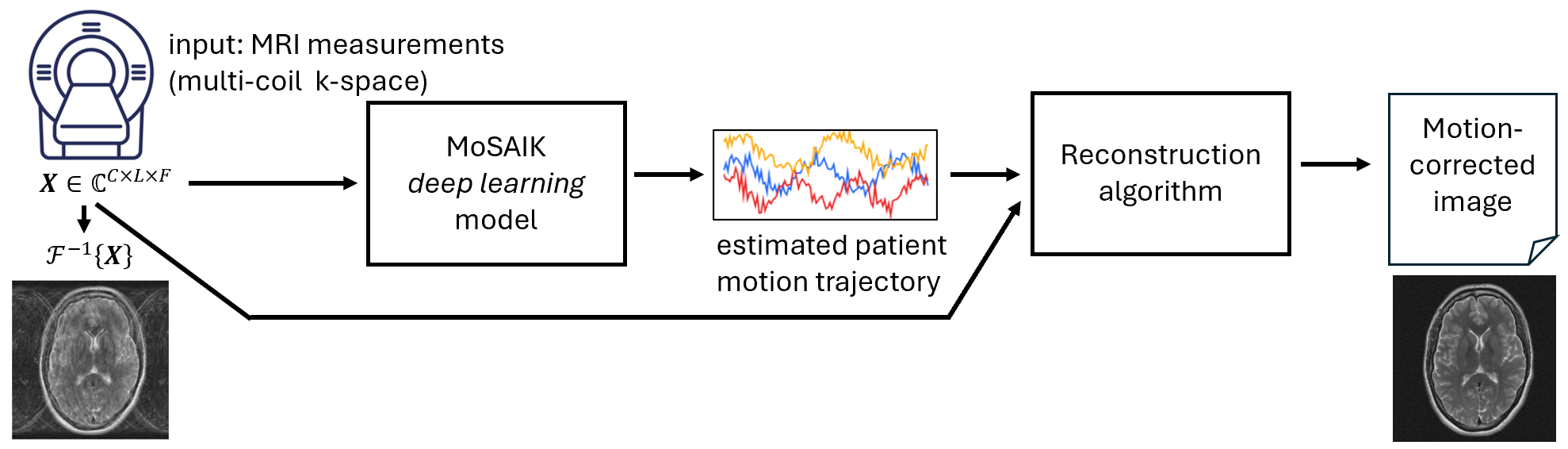}}
\caption{ Overview of the MoSAIK pipeline. Raw scanner measurements ($\bm{X}$) are analysed by our deep neural network to estimate patient's head motion trajectory. A separate non-deep learning algorithm reconstructs the clean image. The two-step approach avoids \textit{deep hallucinations} by design. ($\mathcal{F}^{-1}$ is the inverse Fourier transform).}
\label{mosaik_pipeline}
\end{figure*}



\subsection{Problem Formulation}
The MRI forward model, which describes the underlying physical process of signal formation, can be expressed under idealized conditions by the following equation:
\begin{equation}
    s_c(\boldsymbol{k}) = \int_{FOV} C_c(\boldsymbol{r}) \rho(\boldsymbol{r})e^{-i2\pi  \boldsymbol{k} \cdot \boldsymbol{r}  } \textrm{d}\boldsymbol{r}
\end{equation}
where $\rho \in \mathbb{R}$ is the (weighted) magnetization distribution, $\boldsymbol{k} \in \mathbb{R}^2$ are the k-space coordinates, $\boldsymbol{r} \in \mathbb{R}^2$ the spatial coordinates, $C_c : \mathbb{R}^2 \to \mathbb{C}$ the $c$th coil sensitivity map and $i^2 = -1$.

\begin{figure}
\centerline{\includegraphics[width=4.55in]{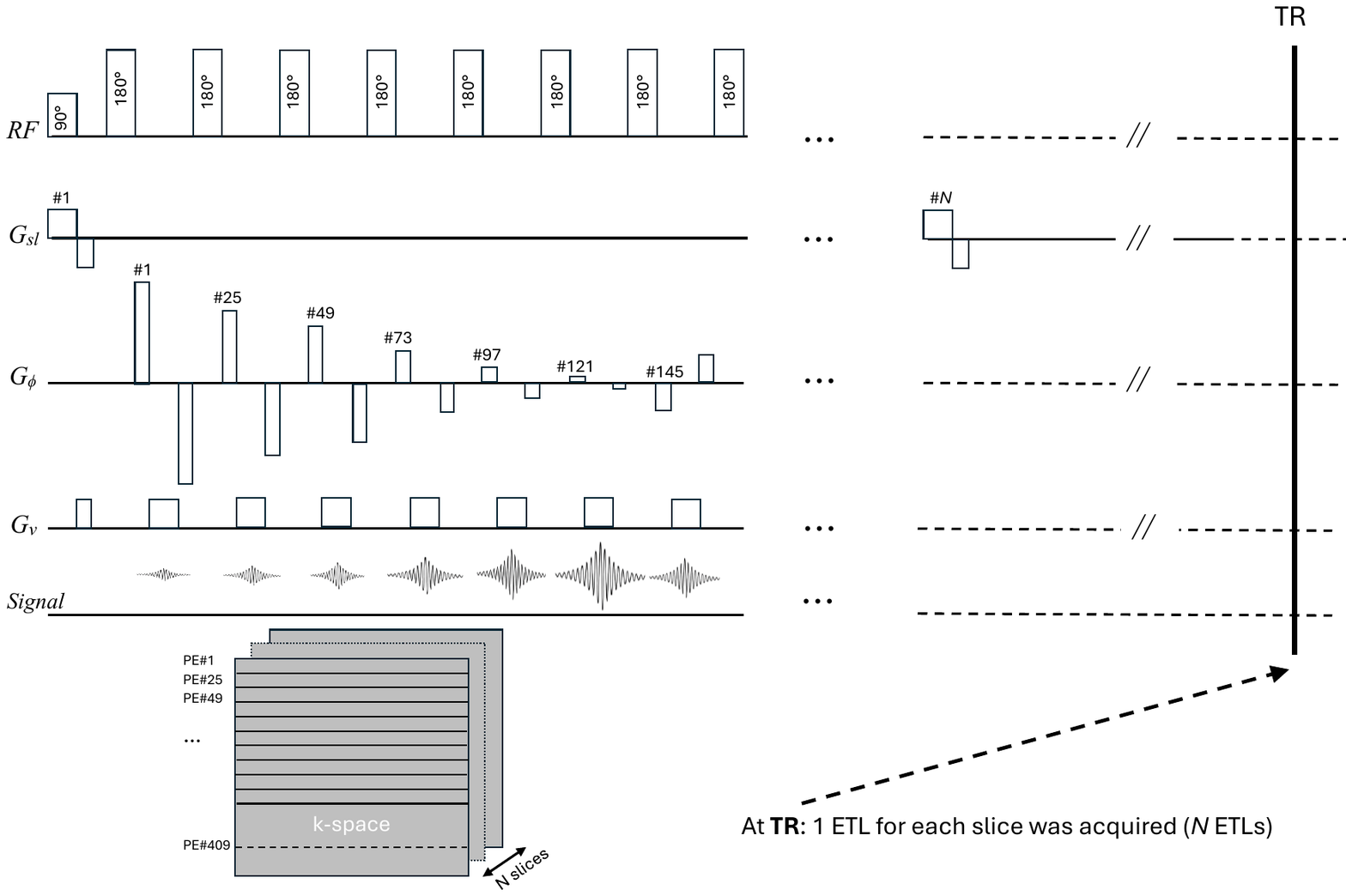}}
\caption{Conceptual description of the TSE pulse sequence we are targeting. During each TR, an ETL is acquired for each slice. The acquired phase encoding lines in each ETL follow a comb-like pattern that covers a wide range of spatial frequencies. Depending the subject's motion trajectory during the scan, slices may not all suffer from the same motion artefacts. A typical ETL ranges from 14 to 18 with $N \approx 33$ slices. }
\label{tse_pulse_seq}
\end{figure}

\begin{figure}
\centerline{\includegraphics[width=4.2in]{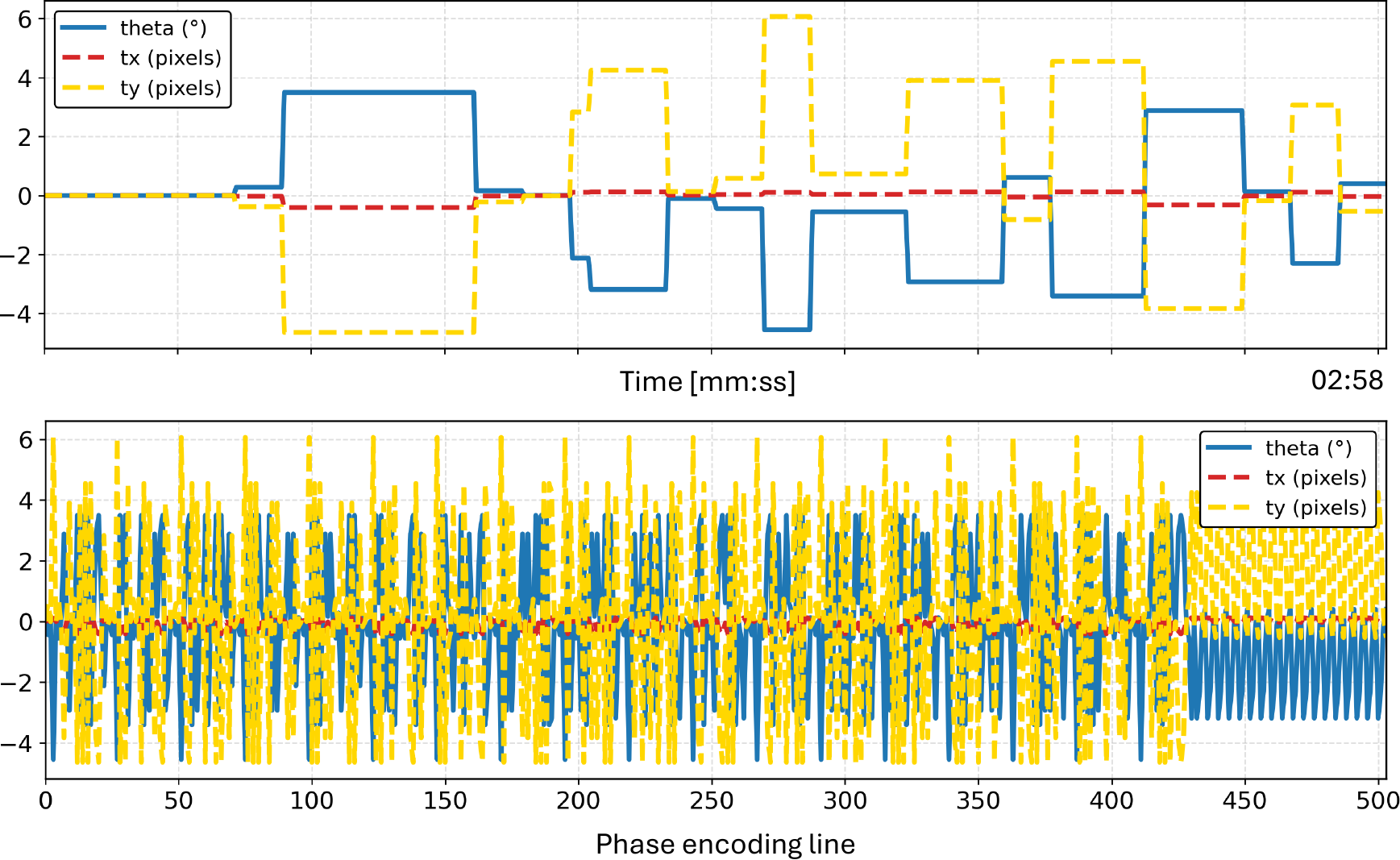}}
\caption{ (Top) Simulated motion trajectory over the full TSE acquisition, in temporal order. (Bottom) One-slice trajectory ordered by acquired PE line, i.e. in k-space order. The oscillation effect is due to the TSE sampling pattern.}
\label{tse_mo_traj}
\end{figure}

Subject motion during k-space acquisition causes inconsistencies across phase encoding (PE) lines, resulting in artefacts such as ghosting or blurring when the image is reconstructed by conventional inverse Fourier transform. Motion is assumed to occur between PE lines, as frequency encoding (FE) is much faster than typical head motion, making motion during FE negligible. As commonly done in related work \cite{dabrowski2024sismik, hossbach2023deep, singh2024data}, we adopt a 2D in-plane rigid-body model to describe head motion. Following signal acquisition on a digital computer, we can use the matrix formalism proposed by Batchelor \textit{et al.} \cite{batchelor2005matrix}, to directly incorporate a motion matrix $\boldsymbol{M}$ parametrized by a motion parameter vector $\boldsymbol{\psi}^{(t)} = [\theta^{(t)}, t_x^{(t)},t_y^{(t)}]$, at each time step $t$, acting on pixels of the underlying vectorized image $\boldsymbol{x} \in \mathbb{C}^{{N_{PE}} \times N_{FE}}$ that we aim to recover :

\begin{equation}
    \boldsymbol{y} = \sum_{t=1}^{N_{PE}} \boldsymbol{P}^{(t)} \boldsymbol{F}\boldsymbol{C}\boldsymbol{M}_{\boldsymbol{\psi}^{(t)}}^{(t)}\boldsymbol{x}
    \label{eq_batchelor}
\end{equation}

where $\boldsymbol{y} \in \mathbb{C}^{{N_{PE}} \times N_{FE}}$ represents the motion-corrupted k-space samples of a given 2D slice, $\boldsymbol{F}$ is the Fourier transform operator, $\boldsymbol{C} \in \mathbb{C}^{{N_c} \times {N_{PE}} \times N_{FE}}$ the coil sensitivity maps and $\boldsymbol{P}^{(t)}$ is the sampling (binary masking) operator which extracts an echo train (ETL) at each time step $t$. For more details on the motion operator $\boldsymbol{M}$ and its relation with head rotation center and resulting residual translations, please refer to \cite{dabrowski2024sismik}, where we present a concrete description of the 2D in-plane rigid-body model for head motion. 
As shown in \autoref{tse_pulse_seq}, each time step $t$ in \autoref{eq_batchelor} corresponds to the acquisition of a single ETL, which samples a broad range of spatial frequencies (except perhaps the final one, which has to fill in the remaining gaps in k-space). Slices should be considered individually since motion can occur at any time and possibly multiple times during one TR. In a given slice, each ETL can possibly be corrupted with different motion parameters depending on patient motion during the scan, resulting in a complex motion trajectory when viewed as a function of PE lines as shown in \autoref{tse_mo_traj}. 

We split the ill-posed inverse problem of recovering $\boldsymbol{x}$ from the motion-corrupted measurements $\boldsymbol{y}$ in two steps.
\begin{enumerate}
    \item From $\boldsymbol{y}$, estimate motion parameters $\boldsymbol{\psi}$ at each time step $t$, for each 2D slice, using a trained deep neural network.
    \item From the estimates in (1), reconstruct a restored version of $\boldsymbol{x}$ that is by construction free of any deep hallucination artefacts.
\end{enumerate}

\subsection{Input representation and normalization}
Working in the k-space domain is challenging due to its high dynamic range. In our previous work, normalization was performed using mean and standard deviation maps computed on the whole training set. This is inconvenient, since it creates a strong coupling with the training data which may negatively impact generalization. We found that using a log-polar representation, i.e., the logarithm of the k-space magnitude combined with the unmodified phase facilitates learning. Applying a conventional z-score normalization to each channel individually further improves training performance.

\subsubsection{K-space input cropping}

K-space was cropped along the frequency-encoding (FE) direction to reduce the input dimensionality of MoSAIK. Crop sizes ranging from 64 to 256 FE samples, centered around the DC component, were evaluated. The FE crop size was treated as a key hyperparameter, as it determines the range of FE spatial frequencies available to the network and strongly affects training and validation performance. The best performance was obtained with 192 FE samples.

\subsubsection{Coil selection algorithm}\label{coilselalgo}

To reduce the input size to the neural network, we selected a subset of four coils from the original multi-coil k-space data. The number of available coils per acquisition ranged from 16 to 64. Our selection was based on two criteria: high SNR ratio and low inter-coil correlation. Specifically, all $N_c$ coils were first sorted in descending order of SNR using a wavelet transform, and the top 16 were retained. Among these, we computed the pairwise correlations and applied a greedy selection algorithm, starting from the highest-SNR coil, to identify four coils with minimal mutual correlation. We avoided the use of virtual coil algorithms in order to preserve the raw data.



\subsection{MoSAIK deep learning model - motion parameter estimation}
We have previously shown that motion parameters can be estimated in a relative manner using CNNs applied to adjacent PE lines from classical 2D Spin-Echo brain acquisitions \cite{dabrowski2024sismik}. The k-space sampling pattern in TSE is such that local convolutions are not well suited for capturing relationships between multiple ETLs (although deeper networks can potentially mitigate this problem by increasing the effective receptive field). The attention mechanism inherent to the Transformer architecture was chosen to address this need of global k-space analysis.

\subsubsection{Network architecture}
The proposed approach called MoSAIK which stands for Motion Search with Artificial Intelligence in K-space, leverages multi-head Transformer \textit{encoder} modules with self-attention to extract motion parameters from embedded raw multi-coil motion-corrupted k-space measurements. We hypothesize that motion estimation benefits from self-attention, which enables modeling global dependencies across k-space through bidirectional attention over the full input (unlike autoregressive decoders that rely on causal masking and sequential inference).

The input to the network consists of raw multi-coil complex k-space data. The complex representation of each coil data can be either log-polar (logarithm of magnitude and phase) or Cartesian (real and imaginary).
Since the number of coils per slice ranges from 16 to 64, we reduce the input dimensionality by applying a custom coil selection algorithm (described in \autoref{coilselalgo}), which retains 4 representative coils per slice.

\begin{figure*}[t]
    \centering
    \includegraphics[width=\textwidth]{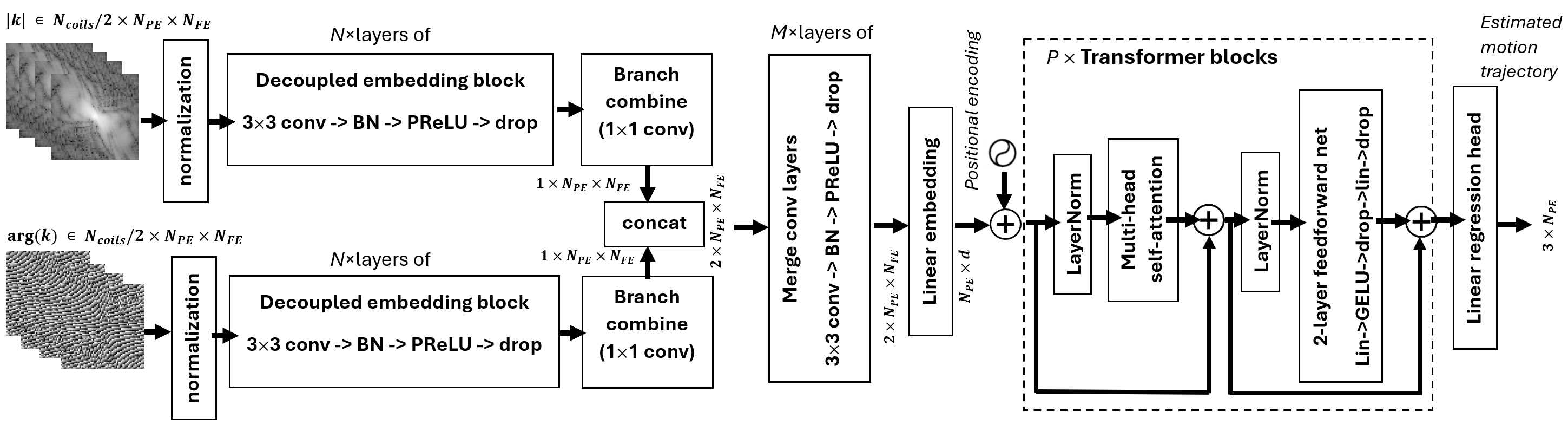}
    \caption{MoSAIK deep learning model. Input is split in $2\times 4$ channels grouped by representation (real and imaginary or magnitude and phase) and processed in a decoupled manner in parallel. Following this convolutional embedding subnetwork, the extracted representation is fed to a Transformer encoder network, operating in dimension $d$, and finally to a linear regression head which predicts 3 motion parameters per PE line.}
    \label{MoSAIK_arch}
\end{figure*}

We posit that a rich embedding mechanism plays a central role by providing feature extraction and a form of input data compression. Therefore, we propose an \textit{Decoupled Representation Merging} (DRM) convolutional embedding sub-network  that pre-processes the input before feeding a compressed representation to the transformer module. The DRM splits the complex input into log-magnitude and phase or real and imaginary, depending on the chosen data representation, and applies a sequence of convolutional layers (conv-batchnorm-PReLU-dropout blocks) in parallel to both representations. The input shape is preserved in both branches and finally merged by a ''conv-merging'' module to a shape of 2 channels (log-magn. and phase or real and imaginary) but same width and height. The latter is linearly projected to the embedding dimension of the Transformer network and the forward pass proceeds through $N$ standard Transformer layers. The $N_{PE} \times 3$ rigid-body motion parameters, i.e., a rotation angle and two translations, per PE line, are finally obtained as the output of a linear regression head added on top of the last Transformer layer. LayerNorm layers are placed before the multi-head self-attention module (''Pre-LN architecture'') and before the multi-layer perceptron, as recommended in \cite{xiong2020layer}, to improve training stability and avoid the ''learning rate warm-up'' phase needed in ''Post-LN Transformers''. A detailed diagram of MoSAIK's architecture is proposed in \autoref{MoSAIK_arch}. 

\subsubsection{Network training}
MoSAIK was trained using the supervised learning paradigm, hence a large labeled training set was needed. Fortunately, it is possible to simulate rigid-body motion from motion-free source acquisitions. To enable realistic simulations that generalize to unknown in vivo motion, we leveraged the physics of the TSE acquisition using scanner metadata extracted along with the raw data, which provides important timing and sequence parameters (e.g., ETL sampling patterns). Building on our previous work with SISMIK for classical 2D Spin-Echo acquisitions, we were confident that motion information could be extracted directly from k-space without requiring a motion-free reference.

The model is implemented in PyTorch and training was performed using the Adam optimizer with mean squared error (MSE) loss between the estimated and ground truth (simulated) motion trajectories. Starting from a learning rate of $3\times10^{-5}$, we used cosine annealing to gradually decrease it throughout training. We used a dropout rate of 0.2 and a batch size of 8. The Transformer embedding dimension was varied between 64 and 256. 

\subsubsection{Inference - motion trajectory estimation}

The network outputs a matrix containing three in-plane rigid-body motion parameters for each k-space phase-encoding line. The resulting motion trajectory can then be used to reverse the effect of motion during image reconstruction.


\subsection{Motion trajectory post-processing}
To improve the robustness of the estimated motion trajectories, a simple post-processing step was applied to the motion trajectories predicted by MoSAIK. Median filtering followed by total variation (TV) denoising was used to remove isolated outliers and high-frequency fluctuations in the estimated trajectories. This correction was performed without using any image-domain information or ground-truth motion trajectory, relying only on the ''weak'' physical prior that head motion is expected to vary continuously over time. In particular, isolated millisecond-scale spikes in the estimated trajectory were considered unlikely to correspond to physically plausible head motion. The resulting trajectories were smoother, and simulations showed that this post-processing step tends to reduce motion-estimation errors.


\subsection{Reconstruction algorithm}

In the reconstruction stage, our deep neural network estimates the motion trajectory, which is used to parameterize the \textit{motion-aware} MRI forward model. The corrected image is then reconstructed from the acquired motion-corrupted k-space data by solving a regularized motion-aware inverse problem. Since we use deep learning for motion parameter estimation only and not directly to infer the corrected image, the reconstruction remains constrained by the acquired measurements through the data-consistency term, thereby avoiding \textit{deep hallucinations} by design.

\subsubsection{Theory}
Let
\[
\widehat{\bm{\psi}}^{(\ell)}
=
\begin{bmatrix}
\widehat{\theta}_{\ell} &
\widehat{t}_{x,\ell} &
\widehat{t}_{y,\ell}
\end{bmatrix}^{\top}
\in \mathbb{R}^{3}
\]
denote the estimated in-plane rigid-body motion parameters for phase-encoding line \(\ell\), where \(\widehat{\theta}_{\ell}\) is the in-plane rotation angle and \(\widehat{t}_{x,\ell}\) and \(\widehat{t}_{y,\ell}\) are the translations. The full estimated trajectory is written as
\[
\widehat{\bm{\Psi}}
=
\begin{bmatrix}
\widehat{\bm{\psi}}^{(1)} &
\widehat{\bm{\psi}}^{(2)} &
\cdots &
\widehat{\bm{\psi}}^{(N_{\mathrm{PE}})}
\end{bmatrix}^{\top}
\in \mathbb{R}^{N_{\mathrm{PE}} \times 3},
\]
where \(N_{\mathrm{PE}}\) is the number of acquired phase-encoding lines.

Let \(\bm{x} \in \mathbb{C}^{N}\) be the unknown motion-free complex image, vectorized over the \(N\) image voxels, and let \(\bm{y}_{\mathrm{corr}} \in \mathbb{C}^{M}\) be the measured motion-corrupted multi-coil k-space data, vectorized over all acquired k-space samples and receiver coils. The motion-aware MRI forward model is written as
\[
\bm{y}_{\mathrm{corr}}
=
\bm{A}(\widehat{\bm{\Psi}})\bm{x}
+
\bm{\varepsilon},
\label{eq_motion_forward_model}
\]
where \(\bm{\varepsilon} \in \mathbb{C}^{M}\) models the measurement noise. The motion-aware encoding operator \(\bm{A}(\widehat{\bm{\Psi}})\) is defined by 
\[
\bm{A}(\widehat{\bm{\Psi}})
=
\begin{bmatrix}
\bm{P}^{(1)} \bm{F} \bm{C} \bm{M}_{\widehat{\bm{\psi}}^{(1)}} \\
\bm{P}^{(2)} \bm{F} \bm{C} \bm{M}_{\widehat{\bm{\psi}}^{(2)}} \\
\vdots \\
\bm{P}^{(N_{\mathrm{PE}})} \bm{F} \bm{C} \bm{M}_{\widehat{\bm{\psi}}^{(N_{\mathrm{PE}})}}
\end{bmatrix}.
\label{eq_motion_encoding_operator}
\]
Here, \(\bm{M}_{\widehat{\bm{\psi}}^{(\ell)}}\) is the image-domain rigid-body motion operator associated with the estimated motion state of line \(\ell\), \(\bm{C}\) is the multi-coil sensitivity encoding operator, \(\bm{F}\) is the Fourier encoding operator, and \(\bm{P}^{(\ell)}\) selects the k-space samples acquired at phase-encoding line \(\ell\). The total number of measurements is \(M = N_{\mathrm{c}} N_{\mathrm{samp}}\), where \(N_{\mathrm{c}}\) is the number of receiver coils and \(N_{\mathrm{samp}}\) is the total number of acquired k-space samples per coil.

The motion-corrected image was reconstructed by solving the regularized inverse problem
\[
\widehat{\bm{x}}_{\mathrm{clean}}
=
\arg\min_{\bm{x}}
\frac{1}{2}
\left\|
\bm{A}(\widehat{\bm{\Psi}})\bm{x}
-
\bm{y}_{\mathrm{corr}}
\right\|_{2}^{2}
+
\lambda \mathcal{R}(\bm{x}),
\label{eq_motion_inverse_problem}
\]
where \(\widehat{\bm{x}}_{\mathrm{clean}}\) is the reconstructed motion-corrected image, \(\mathcal{R}\) is the regularization penalty, and \(\lambda > 0\) controls the regularization strength. The data-fidelity term enforces agreement between the measured k-space data and the data predicted by the motion-aware forward model.


Two regularized reconstruction strategies were used. The first one was formulated as a CG-SENSE reconstruction with quadratic Tikhonov regularization:
\[
\mathcal{R}_{\mathrm{Tik}}(\boldsymbol{x})
=
\frac{1}{2}
\left\|
\boldsymbol{\Gamma}\boldsymbol{x}
\right\|_{2}^{2},
\label{eq_tikhonov_regularization}
\]
where \(\boldsymbol{\Gamma}\) is the Tikhonov regularization operator. In the simplest case, \(\boldsymbol{\Gamma}\) is the identity matrix, which favors the minimum-energy solution among data-consistent images and tends to suppress components lying in the null space of \(\boldsymbol{A}\). The resulting optimization problem is quadratic and leads to the normal equations
\[
\left[
\boldsymbol{A}(\widehat{\boldsymbol{\Psi}})^{\mathrm{H}}
\boldsymbol{A}(\widehat{\boldsymbol{\Psi}})
+
\lambda
\boldsymbol{\Gamma}^{\mathrm{H}}\boldsymbol{\Gamma}
\right]
\widehat{\boldsymbol{x}}_{\mathrm{clean}}
=
\boldsymbol{A}(\widehat{\boldsymbol{\Psi}})^{\mathrm{H}}
\boldsymbol{y}_{\mathrm{corr}},
\label{eq_tikhonov_normal_equations}
\]
where \((\cdot)^{\mathrm{H}}\) denotes the Hermitian transpose. These equations were solved iteratively using conjugate gradients, requiring only the application of the forward operator \(\boldsymbol{A}(\widehat{\boldsymbol{\Psi}})\) and its adjoint \(\boldsymbol{A}(\widehat{\boldsymbol{\Psi}})^{\mathrm{H}}\).

The second reconstruction strategy used total variation regularization:
\[
\mathcal{R}_{\mathrm{TV}}(\boldsymbol{x})
=
\left\|
\nabla \boldsymbol{x}
\right\|_{2,1}
=
\sum_{n=1}^{N}
\sqrt{
\left|
(\boldsymbol{D}_{x}\boldsymbol{x})_{n}
\right|^{2}
+
\left|
(\boldsymbol{D}_{y}\boldsymbol{x})_{n}
\right|^{2}
},
\label{eq_tv_regularization}
\]
where $||.||_{2,1}$ denotes the (2,1)-mixed-norm\cite{baritaux2010priori}. The \(\nabla\) symbol denotes the discrete spatial gradient, defined using finite-difference operators \(\boldsymbol{D}_{x}\) and \(\boldsymbol{D}_{y}\) along the two image axes, and \(n\) indexes the image voxels.
The corresponding TV-regularized inverse problem was solved using iterative first-order optimization methods, namely the alternating direction method of multipliers (ADMM) and proximal gradient descent (PGD). Both solvers require repeated applications of the motion-aware forward operator \(\boldsymbol{A}(\widehat{\boldsymbol{\Psi}})\) and its adjoint \(\boldsymbol{A}(\widehat{\boldsymbol{\Psi}})^{\mathrm{H}}\).


\subsubsection{Practical implementation details}

The conceptual operator \(\boldsymbol{M}_{\widehat{\boldsymbol{\psi}}}\) represents the in-plane rigid motion undergone by the object during acquisition of phase-encoding line \(\ell\), where \(\hat\psi^{(\ell)}=(\hat\theta_\ell,\hat t_{x,\ell},\hat t_{y,\ell})\) is predicted by MoSAIK. For the numerical reconstruction, this motion model was implemented using a NUFFT\footnote{Non-uniform fast Fourier transform.}-based parallel MRI operator. Instead of explicitly resampling the image for each line, the rotational component \(\hat\theta_\ell\) was incorporated by rotating the prescribed TSE Cartesian k-space coordinates of that line, resulting in a non-Cartesian trajectory. Coil sensitivities were included in the forward and adjoint operators through SENSE encoding. The translational parameters were treated separately using the Fourier shift theorem: the measured k-space data was multiplied by conjuguate phase ramps to remove the effects of translations prior to rotational motion-correction. Finally, the rotation-aware multi-coil SENSE NUFFT operator was implemented as a DeepInv LinearPhysics module, with torchkbnufft used for the NUFFT evaluations\cite{tachella2025deepinverse, muckley20tah}.


\subsection{Datasets}
\subsection{In-silico dataset }


Two-dimensional multi-slice T2w Turbo Spin-Echo (TSE) raw k-space data were collected from two clinical 3T MRI scanners: MAGNETOM Prisma fit and MAGNETOM Cima.X (Siemens Healthineers, Erlangen, Germany). The acquisitions were performed using the routine clinical TSE protocol in use at our institution at the time of data collection, with turbo factor 18, acquisition matrix (NPE $\times$ NFE) of (504 $\times$ 512), voxel size 0.4 $\times$ 0.4 $\times$ 4.0 mm$^3$, FOV of approx. 220 $\times$ 220 mm, 30-33 slices, TR = 6090 ms, TE = 108 ms, GRAPPA acceleration $\times$ 2, NSA = 2, and TA = 2 min 58 s. 

Each of the two averages was sampling complementary even and odd phase-encoding lines and hence the acquired raw data contained all phase-encoding lines and were treated as fully sampled. After thorough manual visual screening, 53 motion-free acquisitions, corresponding to approximately 1700 slices from 22 patients and 31 phantoms, were retained as source data for simulation.

From this curated source dataset, we generated a large-scale data augmentation set of $10^6$ synthetic motion-corrupted examples. This augmentation was grounded in the physics of MRI acquisition and in the phase-encoding timings of the TSE sequence, rather than relying on arbitrary k-space perturbations or random line replacement. Simulated motion was applied according to the mathematical 2D rigid-body motion model, which is well suited to brain MRI because head motion is commonly approximated by rotations and translations. Each simulation contained 1 to 32 motion events. Rotation angles ($\theta$) were sampled from six zero-mean Gaussian distributions with standard deviations ranging from $0.25^\circ$ to $2.5^\circ$, chosen to represent different levels of motion severity across patient populations. The associated translations ($t_x$ and $t_y$) were derived from the same rigid-body motion model, with the center of rotation randomly sampled near the image center. The aim was not to reproduce exact patient-specific motion trajectories, but rather to generate a broad and physically consistent set of relative k-space line discrepancies for training MoSAIK over the targeted range of motion severity.

\subsubsection{In-vivo datasets}

An independent in vivo test set was acquired using the Choreography-controlled (ChoCo) protocol presented in \cite{dabrowski2022choreography}. Five volunteers were recruited and instructed to perform predefined head-motion choreographies, resulting in real motion artefacts of varying severity. The anatomical scans were acquired using the same TSE protocol described above.

\subsubsection{Dataset splits}
Among the 53 motion-free acquisitions, approximately 10\% were set aside for validation, while the remaining 90\% were used for training. The in vivo ChoCo test set was never used for neither training nor validation.


\subsection{Experimental setup}

\subsubsection{Hardware setup}
Motion simulations and neural network training and inference were performed in part on an HPC cluster and on a high-range server equipped with NVIDIA Blackwell GPUs with 96 GB of VRAM each. Training with a million simulated examples with the proposed deep neural network architecture (10 million parameters) for 50 epochs took $\sim$ 8 days.

\subsubsection{Ablation tests}

Ablation tests were performed to evaluate the contribution of the different neural network architectural components. We tested different configurations of the convolutional embedding module, varying the number of convolutional layers and filters, to assess its impact on motion-estimation performance.
We then removed this module entirely, yielding a transformer-only model. Finally, a purely convolutional network (CNNs) baseline was evaluated to assess performance without the transformer backbone. Preliminary experiments also included recurrent neural network (RNN) baselines.



\subsection{Reconstruction quality evaluation}

\subsubsection{Motion estimation performance}

Motion-estimation performance was evaluated on a simulated validation set, i.e., the target trajectory is known for each example. Predictions were compared with the corresponding target trajectories using mean squared error (MSE), mean absolute error (MAE), and Pearson correlation coefficient. MSE and MAE measure prediction error, while correlation measures trajectory agreement.

For in vivo data, no ground-truth trajectory was available. Therefore, motion estimates were assessed only qualitatively and through the corrected reconstructions.

\subsubsection{Motion correction performance}

\begin{figure*}[!t]
    \centering

    \begin{subfigure}[t]{0.475\textwidth}
        \centering
        \includegraphics[width=\linewidth]{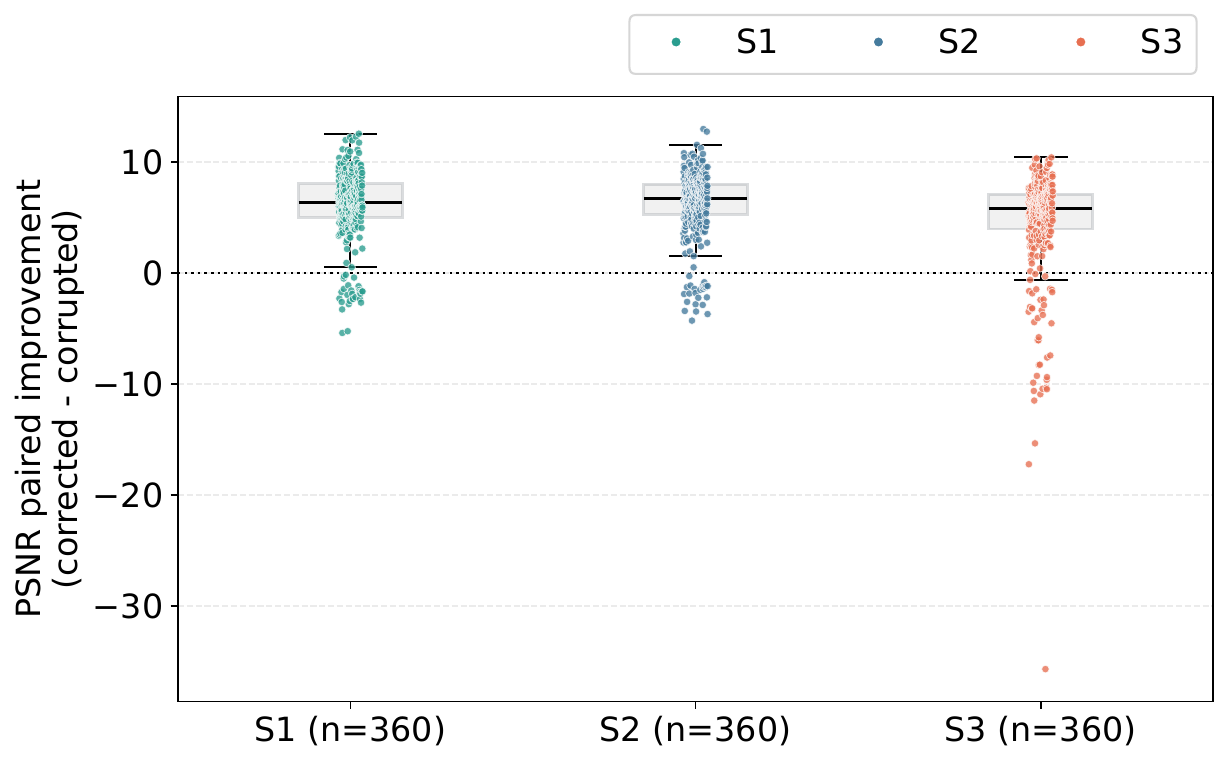}
        \caption{PSNR [dB]}
        \label{fig:simu_gain_psnr}
    \end{subfigure}
    \hfill
    \begin{subfigure}[t]{0.475\textwidth}
        \centering
        \includegraphics[width=\linewidth]{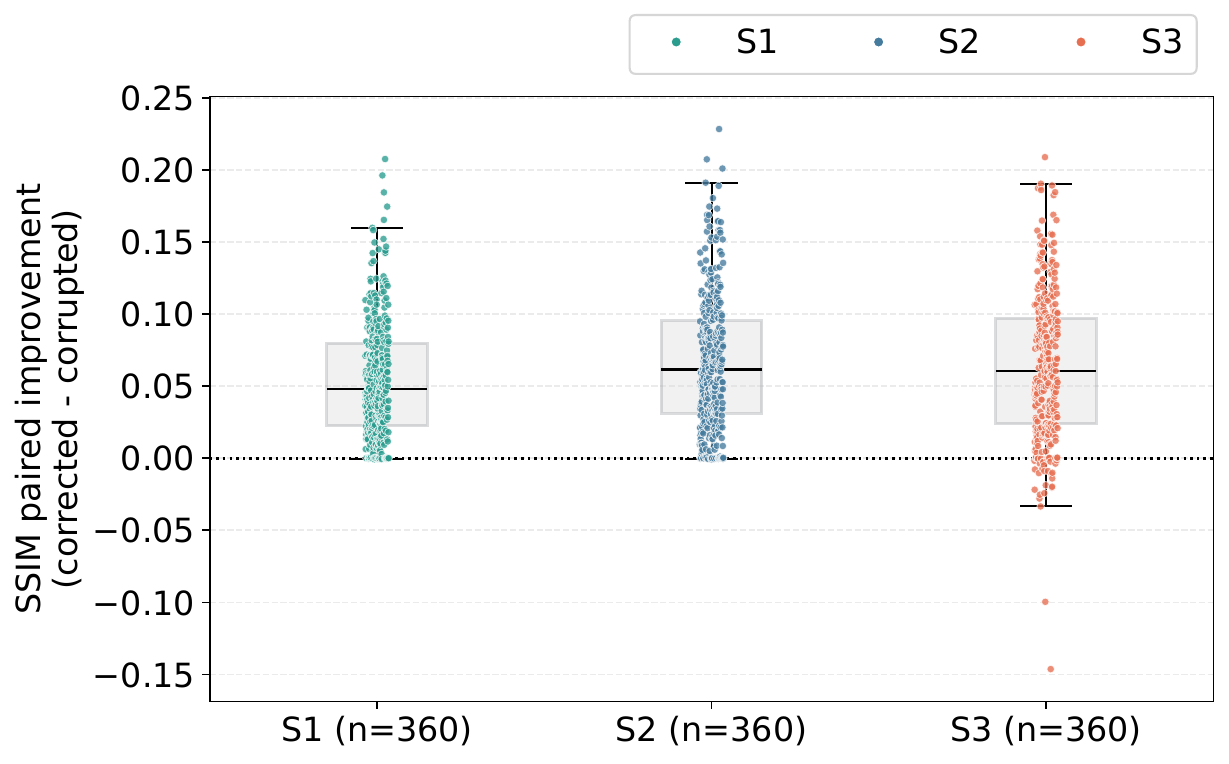}
        \caption{SSIM}
        \label{fig:simu_gain_ssim}
    \end{subfigure}
    \hfill
    \begin{subfigure}[t]{0.475\textwidth}
        \centering
        \includegraphics[width=\linewidth]{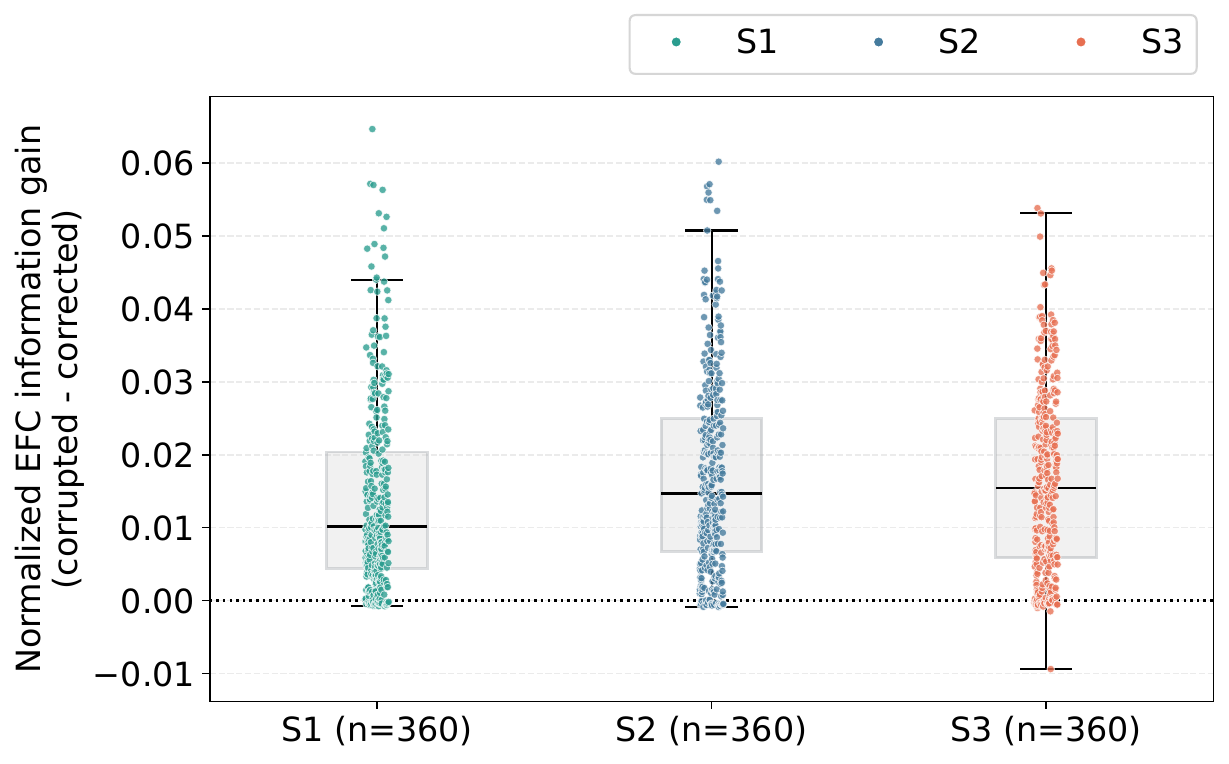}
        \caption{EFC}
        \label{fig:simu_gain_efc}
    \end{subfigure}

    \caption{Subject-wise metric gains on a dataset of \textit{simulated} motion from three subjects unseen during training. Motion was corrected using MoSAIK-estimated trajectories followed by CG-SENSE reconstruction. EFC: Entropy focus criterion. (PSNR is reported for reference only, as it is a poor proxy of perceptual image quality.)}
    \label{fig:simu_subjects_gain}
\end{figure*}

\begin{figure*}[!t]
    \centering

    \begin{subfigure}[t]{0.45\textwidth}
        \centering
        \includegraphics[width=\linewidth]{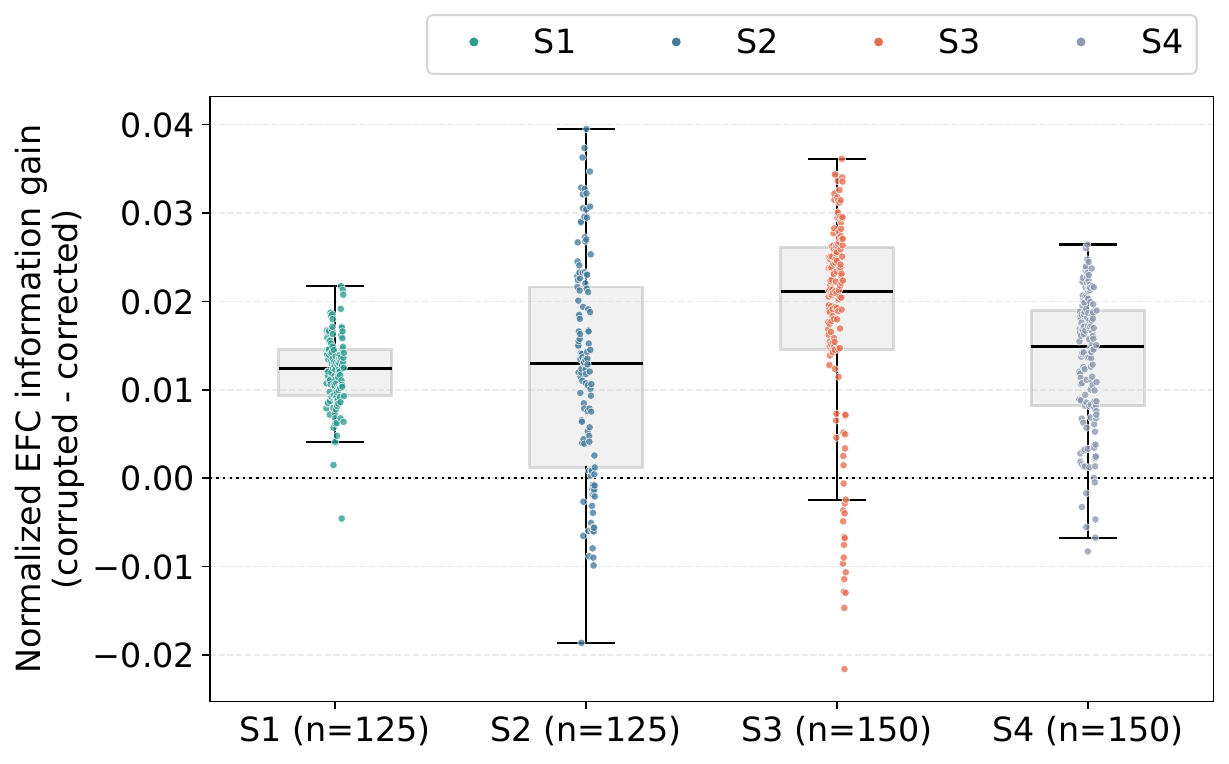}
        \caption{EFC gain by subject}
        \label{fig:efc_by_subj}
    \end{subfigure}
    \hfill
    \begin{subfigure}[t]{0.54\textwidth}
        \centering
        \includegraphics[width=\linewidth]{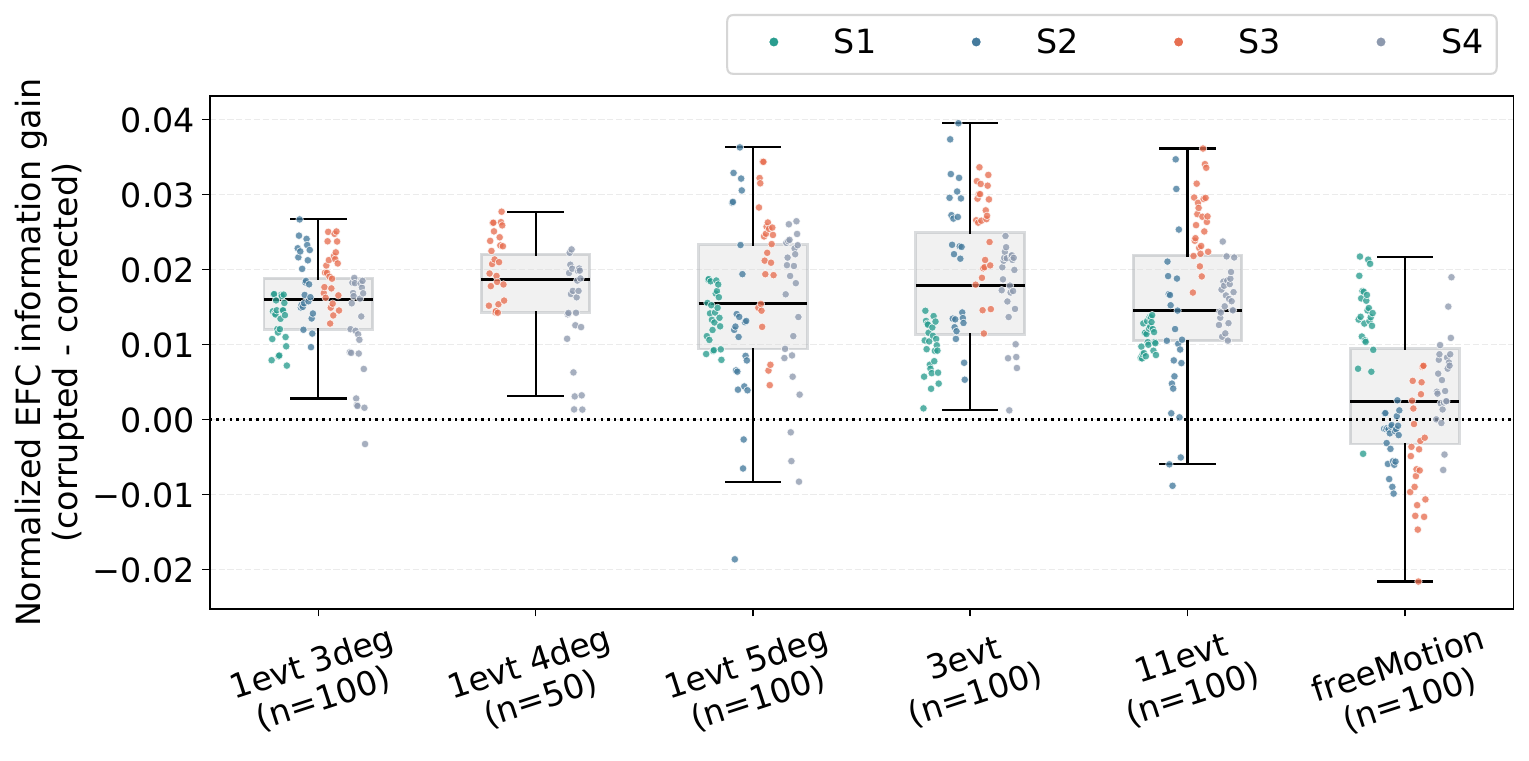}
        \caption{EFC gain by motion type}
        \label{fig:efc_by_type}
    \end{subfigure}
    \hfill

    \caption{EFC gain for four subjects acquired with the \textit{ in-vivo } ChoCo protocol, with six prescribed motion types. Motion correction was performed with motion-aware CG-SENSE reconstruction using MoSAIK-estimated trajectories. EFC gains are summarized by subject (a) and by motion type (b).}
    \label{fig:sinvivo_efc_gain}
\end{figure*}

Reconstructed image quality was assessed using both referenceless and reference-based image-quality metrics. Although EFC does not require a ground-truth image and is hence ''referenceless'', it is essentially a relative metric and is therefore most meaningful when comparing images of the same object. This condition is satisfied in our case, as each comparison involves the motion-corrupted and motion-corrected versions of the same slice.  We report EFC gain in \autoref{fig:simu_subjects_gain} and \autoref{fig:sinvivo_efc_gain} , which is the entropy difference between motion-corrupted and motion-corrected images: higher gain corresponds to improved image quality.

For simulations, where an exact ground-truth reference is available, peak signal-to-noise ratio (PSNR) and structural similarity index measure (SSIM) were also computed.



\subsubsection{Brain masks}
Brain masks were computed using morphological operators and thresholding to restrict the EFC computation to brain tissue and prevent the background from influencing the metric. For each acquisition, the mask was derived from the corresponding motion-free reference image, rather than from the motion-corrupted or corrected images, and was then applied consistently when computing EFC gains.

\section{Results}

\subsection{Results on simulated motion-corrupted data}

MoSAIK was first evaluated on simulated motion-corrupted brain MRI data from subjects unseen during training. The estimated motion parameters were compared against the known ground-truth in-plane rotation $\theta$ and translations $T_x$ and $T_y$. Overall, the predicted motion followed the identity line closely, indicating that the model learned a consistent mapping between k-space corruption patterns and the underlying rigid-body motion parameters. The agreement was strongest within the central motion range, approximately between $-5^\circ$ and $5^\circ$ for rotations and between -5 and 5 pixels for translations. Larger deviations were observed toward the extremes of the simulated parameter range, where the mean estimates increasingly departed from the identity line and the variability increased. These results suggest that MoSAIK provides the most reliable motion estimates within the moderate-motion regime, while performance becomes less stable for larger motion amplitudes.

The histogram overlays show the distribution of the simulated motion parameters and help interpret these deviations. In particular, the central parameter ranges were more densely represented than the extreme ranges, which partly explains the increased uncertainty observed for larger rotations and translations. Despite this, the model preserved the correct overall trend across the full simulated range for all three motion parameters.

Image-quality metrics computed after motion correction further confirmed the benefit of the proposed approach. Across the three simulated test subsets, correction generally led to positive gains in PSNR, SSIM, and entropy focus criterion compared with the corrupted images. The SSIM and entropy-based results showed consistent improvements for most cases, indicating recovery of structural similarity and improved image sharpness/focus after correction. PSNR also improved on average, but displayed a larger number of outliers. This behavior is consistent with the known limitation of PSNR as a purely pixelwise metric that does not always reflect perceptual or structural image quality. Therefore, PSNR was included mainly as a conventional reference metric, whereas SSIM and the entropy focus criterion provide a more informative assessment of the visual and structural quality of the corrected images.

Taken together, these in silico results show that MoSAIK can estimate rigid in-plane motion parameters from motion-corrupted k-space data and that the resulting trajectories lead to measurable image-quality improvements after model-based reconstruction. The results also indicate that the most reliable operating range is the moderate-motion regime, while larger simulated motions remain more challenging and are associated with increased estimation variance.

\begin{figure*}[!t]
\centerline{\includegraphics[width=5.7in]{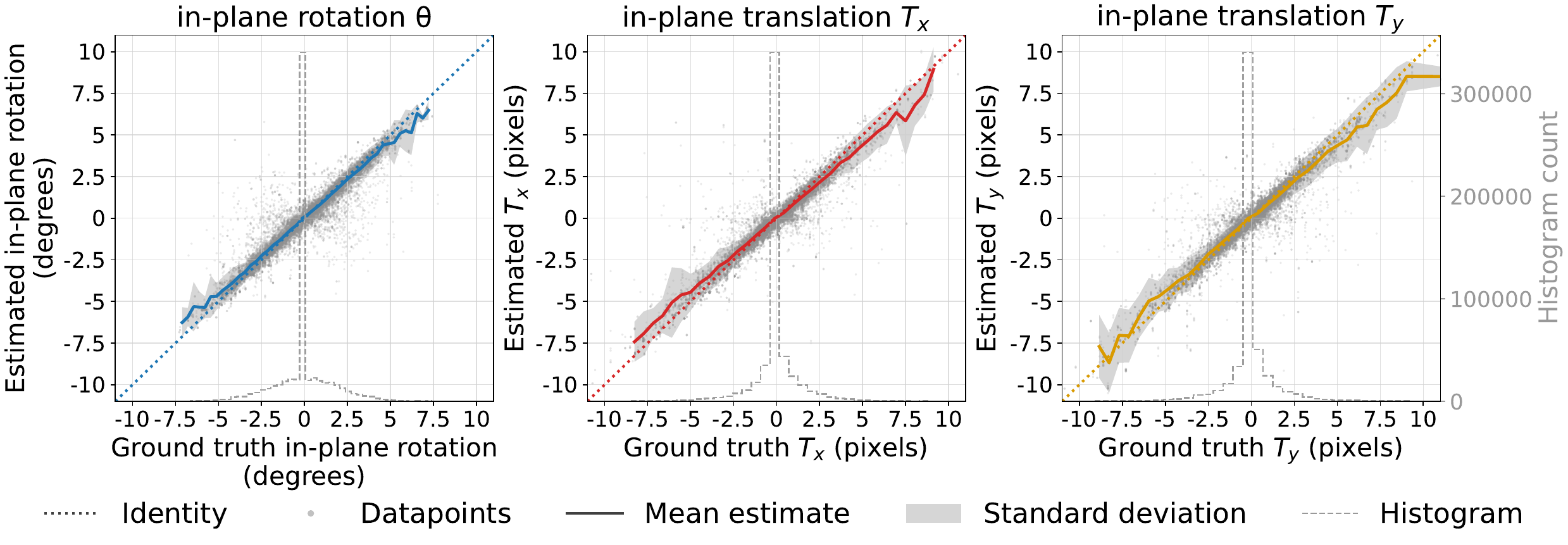}}
\caption{ Aggregate motion estimation on \textit{simulated} motion-corrupted brain MRI from subjects unseen during training. Estimated motion is plotted against ground truth for in-plane rotations \(\theta\), and translations \(T_x\) and \(T_y\). Curves show the average predicted motion for each Ground truth motion range, gray shading shows $\pm$ one standard deviation, and histograms show the simulation parameter distributions.}
\label{res1}
\end{figure*}

\begin{figure*}[!t]
    \centering

    \begin{subfigure}[t]{0.99\textwidth}
        \centering
        \includegraphics[width=\linewidth]{ 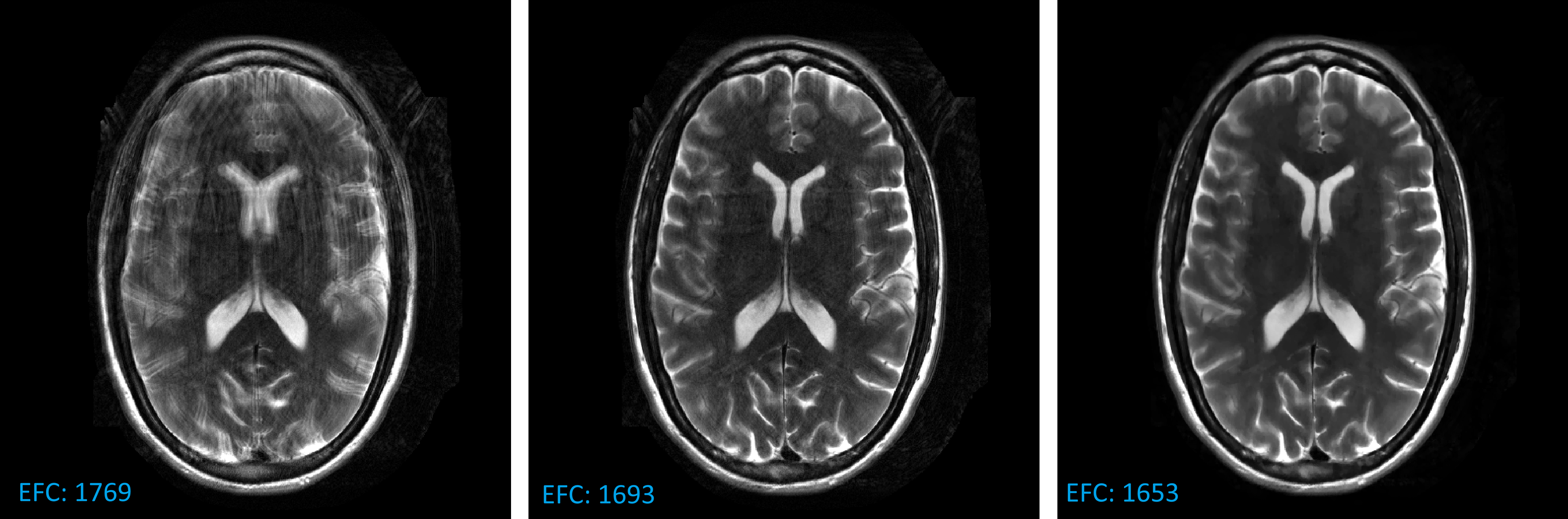 }
        \caption{ In-vivo motion performed by a volunteer following the ChoCo protocol (slice \#9, see appendix for ''pseudo-reference''  \autoref{fig:s3_nomo}). (Left) Inverse Fourier transform of corrupted k-space. (Center) Motion-corrected reconstruction with CG-SENSE and Tikhonov regularization from MoSAIK trajectory in (c). (Right) Same as (Center) but with TV-regularization. (Border effects related to application of a ''brain mask'' computed using morphological operators on the reference motion-free acquisition).  }
        \label{fig:moco_invivo_3evt_1}
    \end{subfigure}
    
      \vspace{2mm}

    \begin{subfigure}[t]{0.999\textwidth}
        \centering
        \includegraphics[width=\linewidth]{ 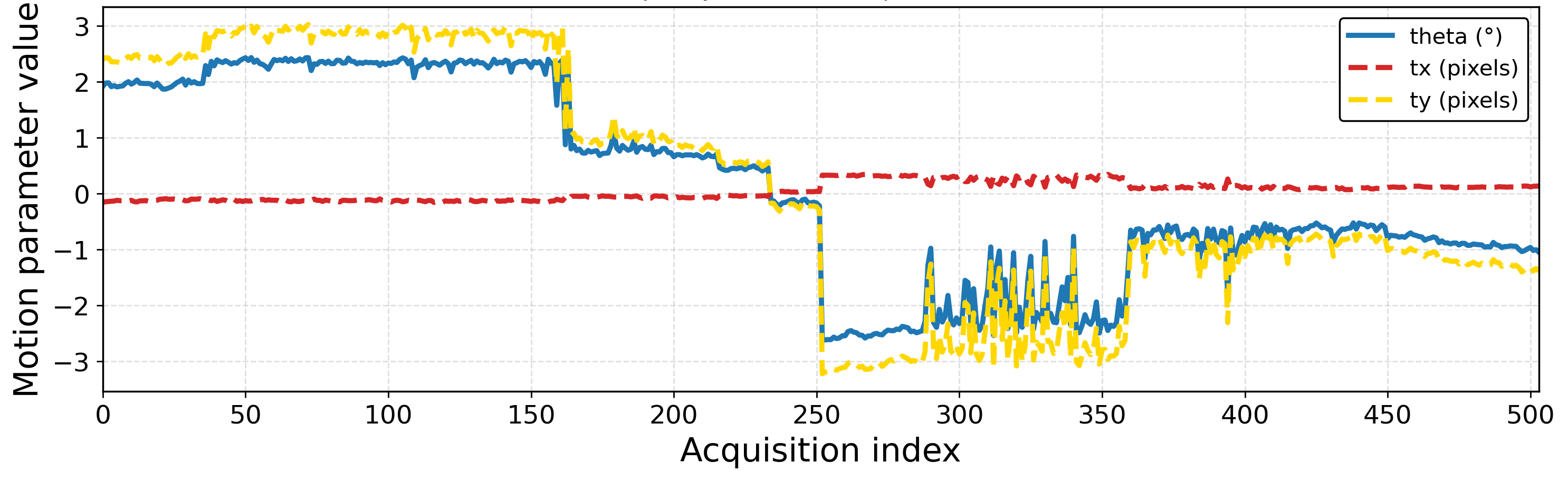  }
        \caption{ Raw MoSAIK-estimated motion trajectory. }
        \label{fig:moco_invivo_3evt_b}
    \end{subfigure}
    
    \vspace{2mm}

        \begin{subfigure}[t]{0.99\textwidth}
        \centering
        \includegraphics[width=\linewidth]{ 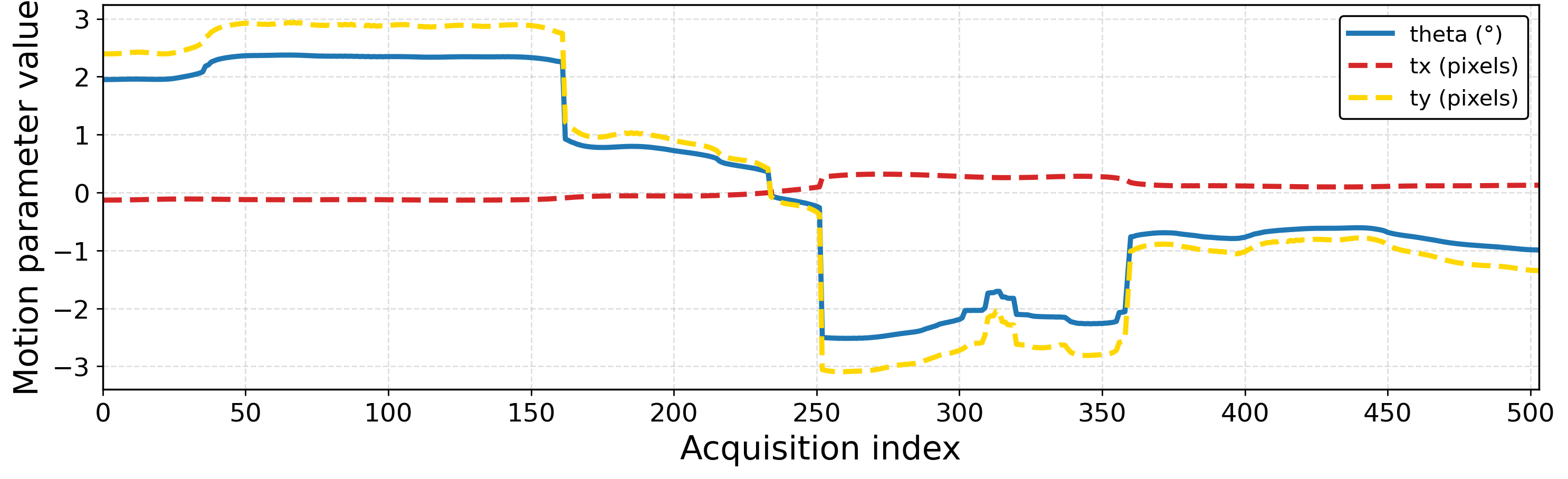  }
        \caption{ Post-processed (median and TV denoising) motion trajectory. }
        \label{fig:moco_invivo_3evt_b}
    \end{subfigure}

    \caption{ Representative in vivo ChoCo acquisition in which the volunteer was asked to perform three motion events. The top panel compares the corrupted inverse-Fourier image with motion-aware CG-SENSE reconstructions using MoSAIK-estimated trajectories and either Tikhonov or TV regularization. The lower panels show the raw and post-processed estimated motion trajectories. }
    \label{invivoS3}
\end{figure*}



\begin{figure*}[!t]
    \centering

    \begin{subfigure}[t]{0.99\textwidth}
        \centering
        \includegraphics[width=\linewidth]{ 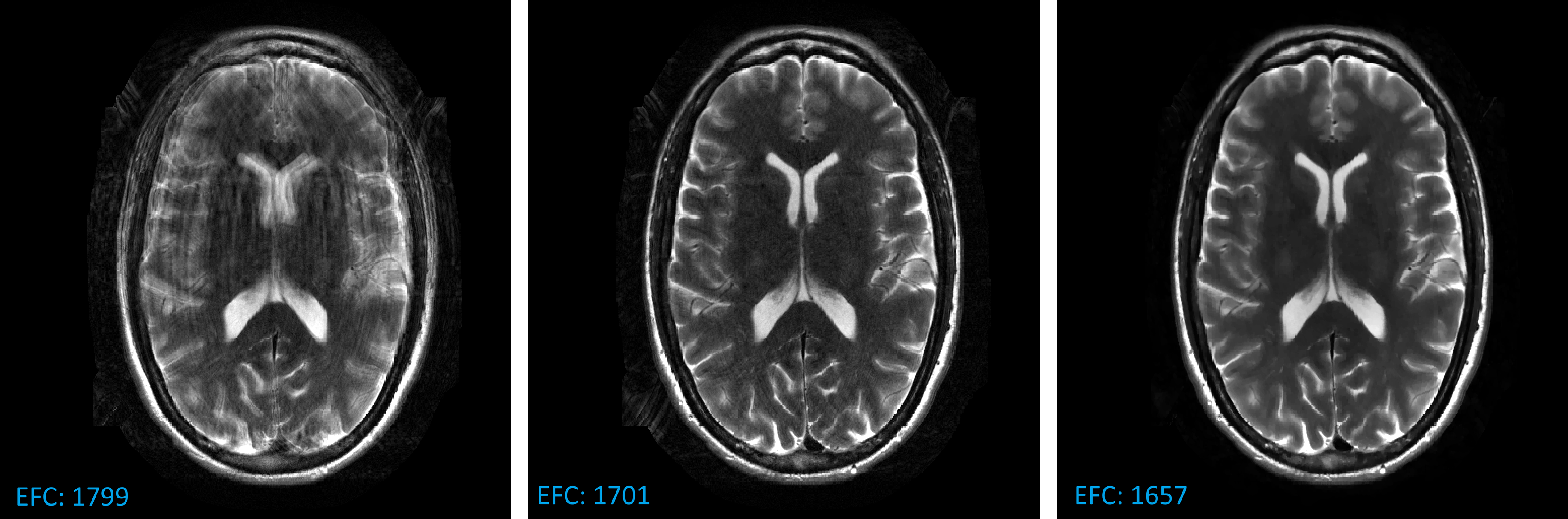 }
        \caption{ Idealized ChoCo trajectory simulation: the prescribed motion pattern (labeled ''GT'' in the trajectory below),requested from the subject, corresponding to the in-vivo trajectory shown in \autoref{fig:moco_invivo_3evt_1}, was applied to the S3 no-motion reference slice shown in \autoref{fig:s3_nomo}. (Center) Motion-corrected from estimated MoSAIK trajectory with CG-SENSE reconstruction and Tikhonov regularization. (c). (Right) Same as (Center) but with TV-regularization.  }
        \label{fig:moco_simu_3evt_1}
    \end{subfigure}
    
      \vspace{2mm}

    \begin{subfigure}[t]{0.999\textwidth}
        \centering
        \includegraphics[width=\linewidth]{  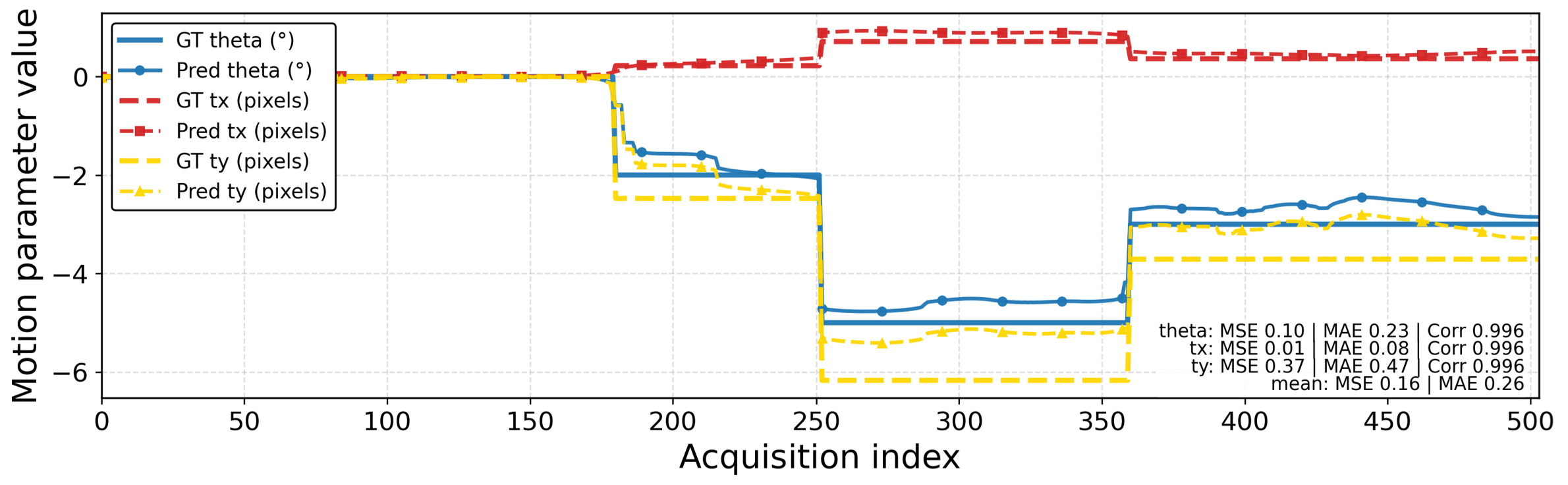 }
        \caption{ MoSAIK-estimated motion trajectory. “GT” denotes the idealized ground-truth motion trajectory used for the simulation, corresponding to the motion pattern requested from S3 and shown in \autoref{fig:moco_invivo_3evt_1}.}
        \label{fig:moco_simu_3evt_b}
    \end{subfigure}

    \caption{ Idealized ChoCo trajectory simulation. A three-event motion trajectory, matching the motion pattern requested during the in-vivo ChoCo acquisition, was applied to the matching S3 no-motion reference slice. (a) Simulated corrupted image and MoSAIK motion-corrected reconstructions. (b) Ground-truth and MoSAIK-estimated motion trajectories. }
    \label{simuS3ChocoIdeal}
\end{figure*}


\subsection{In vivo results}

MoSAIK was further evaluated on in vivo data acquired with the ChoCo protocol, using four volunteers and six prescribed motion conditions. Since ground-truth motion trajectories are not available in vivo, performance was assessed using the entropy focus criterion (EFC) gain between the corrupted and motion-corrected reconstructions. Positive EFC gain indicates an improvement after correction. Please note, that EFC gains were reported for the Tikhonov-regularized reconstructions only. For the TV-regularized reconstructions, the additional $(||\nabla \mathbf{x}||_{2,1})$ penalty promotes spatial smoothness and may therefore affect entropy-based focus metrics independently of motion correction accuracy. These reconstructions were consequently evaluated qualitatively rather than through EFC gain.

Across subjects, motion correction led to consistent positive EFC gains. For all four subjects, the median EFC gain was clearly above zero, and more than 75\% of slices showed a positive gain. This indicates that the MoSAIK-estimated trajectories, when used in the motion-aware CG-SENSE reconstruction, improved image focus for the large majority of in vivo slices.

The analysis by motion type showed a similar trend. For the prescribed single-event and multi-event motion conditions, EFC gains were predominantly positive, with median values clearly above zero and most slices showing improved focus after correction. The free-motion condition remained more challenging: although the median EFC gain was still positive, the distribution was broader and included a larger fraction of negative-gain slices. This is consistent with the less constrained and potentially larger-amplitude nature of free head motion compared with the prescribed ChoCo motion events.

Overall, the in vivo results confirm that MoSAIK-estimated trajectories can provide meaningful image-quality improvements when integrated into a motion-aware CG-SENSE reconstruction. The gains were consistent across subjects and across most prescribed motion conditions, while free motion represented the most difficult case and showed greater variability.

A representative in vivo example is shown in \autoref{invivoS3}. The MoSAIK-estimated trajectory led to a qualitatively improved motion-corrected reconstruction, with reduced motion-related blurring and ringing compared with the corrupted inverse-Fourier image. A brain mask computed from the reference motion-free acquisition was applied during reconstruction, which explains the slightly truncated boundaries visible in some slices. The corresponding simulated experiment in \autoref{simuS3ChocoIdeal} was generated using the ideal ChoCo trajectory prescribed to subject S3. This in silico experiment reproduced artefacts that were qualitatively similar to those observed in vivo, while additionally providing access to the ground-truth trajectory. In this setting, post-processing of the raw MoSAIK trajectory improved the trajectory-estimation metrics and led to visually consistent motion correction, supporting the interpretation of the in vivo S3 result despite the absence of ground-truth motion in vivo. Entropy focus criterion values are displayed (rounded to the nearest integer) for each image to provide a quantitative overview of ghosting and blurring artefact reduction.

\subsection{Ablation tests}

An ablation analysis was performed to assess the contribution of the convolutional embedding module. To reduce training time, models in this analysis were trained on $64 \times 64$ k-space crops. When this module was completely removed, the network achieved near-perfect performance on the training set but completely failed to generalize to unseen data, indicating severe overfitting. Reducing the size of the convolutional embedding preserved generalization, but led to lower performance than the full model. These results suggest that the convolutional embedding module is important for extracting local k-space features and providing a more suitable learned representation before transformer-based motion estimation.

\begin{figure*}[!t]
    \centering

    \begin{subfigure}[t]{0.99\textwidth}
        \centering
        \includegraphics[width=\linewidth]{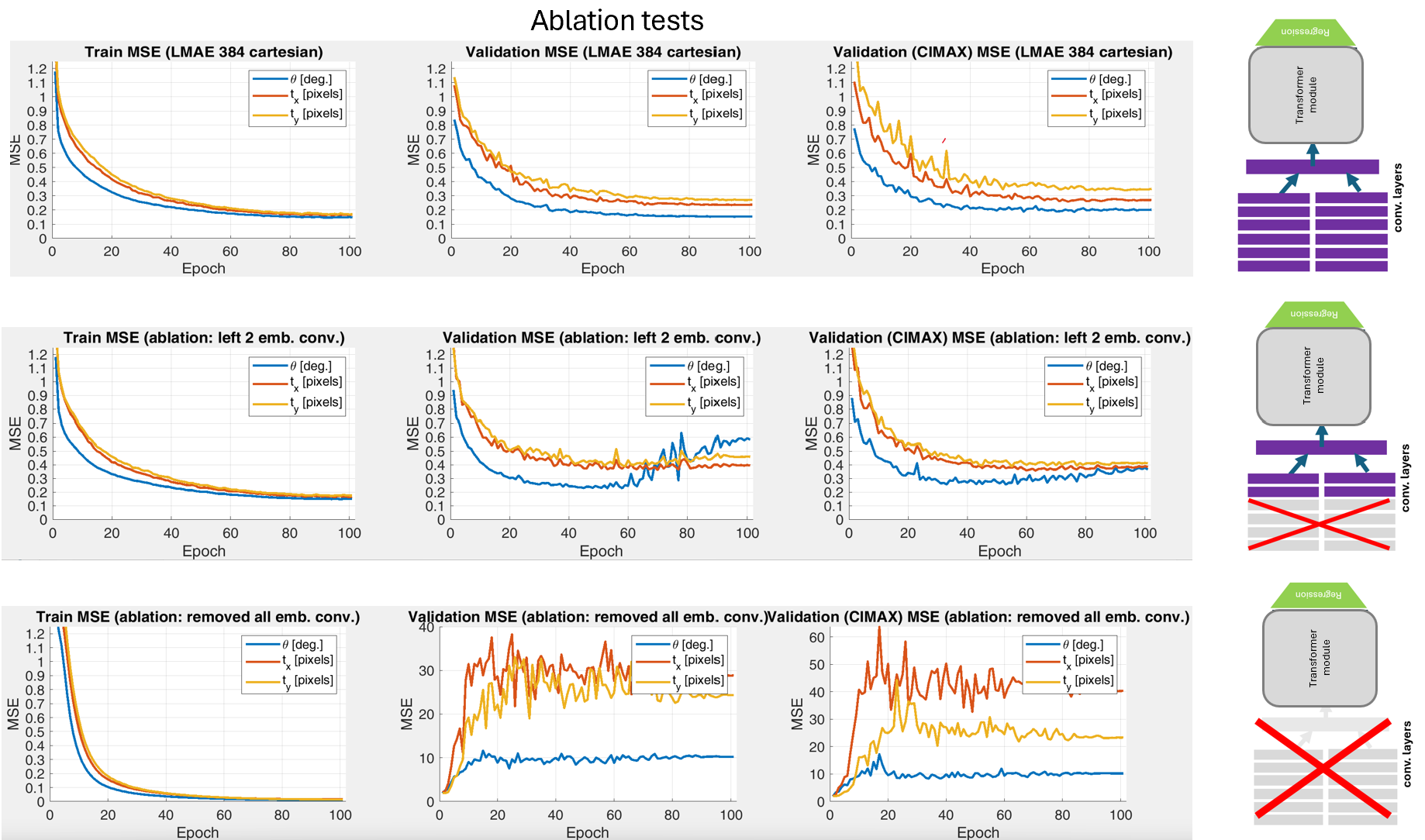}
    \end{subfigure}

    \caption{Ablation tests performed on $64\times64$ cropped k-space dataset showing severe overfitting when convolutional embedding layer is removed. Intermediate generalization performance is exhibited by the model when a small number of convolutional layers is preserved.}
    \label{fig:abl_tests1}
\end{figure*}

Please refer to Appendix \autoref{fig:abl_tests1} for training and validation plots.

\clearpage
\section{Discussion}

This work demonstrates that MoSAIK can estimate rigid in-plane motion trajectories directly from motion-corrupted k-space data and incorporate these estimates into a physics-based reconstruction framework to reduce motion artefacts. A key feature is the separation of motion estimation from image reconstruction: the network predicts motion parameters, which inform the inverse problem used to reconstruct the image. This avoids the risk of hallucinated anatomical structures by design and maintains an explicit link between the acquired data, the estimated motion, and the reconstruction model. The motion-estimation component could also support applications beyond retrospective correction, including motion detection (e.g., in prospective motion-correction settings).

The simulated experiments confirmed that MoSAIK can recover the main motion trends from corrupted k-space data. The agreement between estimated and ground-truth motion was strongest in the moderate-motion range, approximately within $-5^\circ$ to $5^\circ$ for in-plane rotations. This range was deliberately emphasized in the simulation design, since it corresponds to realistic head motion amplitudes expected in a padded head coil. Larger rotations were more challenging and led to larger deviations, which is consistent with the broader extrapolation required outside the most represented training regime.

The in vivo ChoCo experiments further supported the practical relevance of the approach. Since ground-truth motion is not available in vivo, image quality was assessed using EFC gains and qualitative inspection. Positive EFC gains were observed across subjects and across most prescribed motion types, indicating improved image focus after motion-aware reconstruction. The free-motion condition was more variable and showed lower gains, which is likely related to the less constrained nature of this motion. In particular, the estimated trajectories suggested larger rotations than those typically expected in a well-padded clinical acquisition, which may explain the reduced correction performance in this condition.

\begin{figure*}[!t]
    \centering

    \begin{subfigure}[t]{0.95\textwidth}
        \centering
        \includegraphics[width=\linewidth]{ 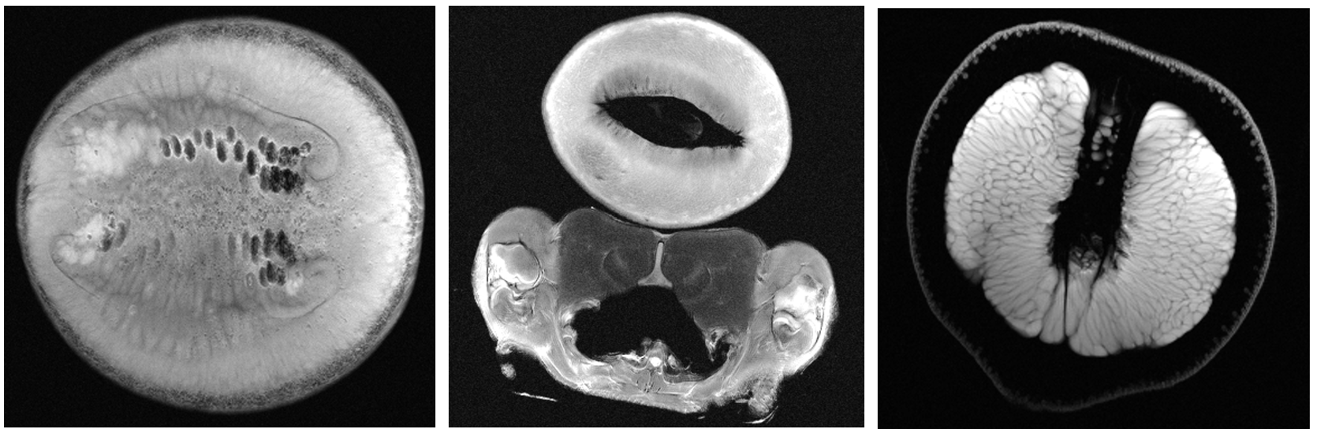 }
        \caption{PhRUITS training set: only MRI scans of fruits, vegetables a raw chicken and 2 stardard water phantoms.}
    \end{subfigure}
    
      \vspace{2mm}

    \begin{subfigure}[t]{0.95\textwidth}
        \centering
        \includegraphics[width=\linewidth]{  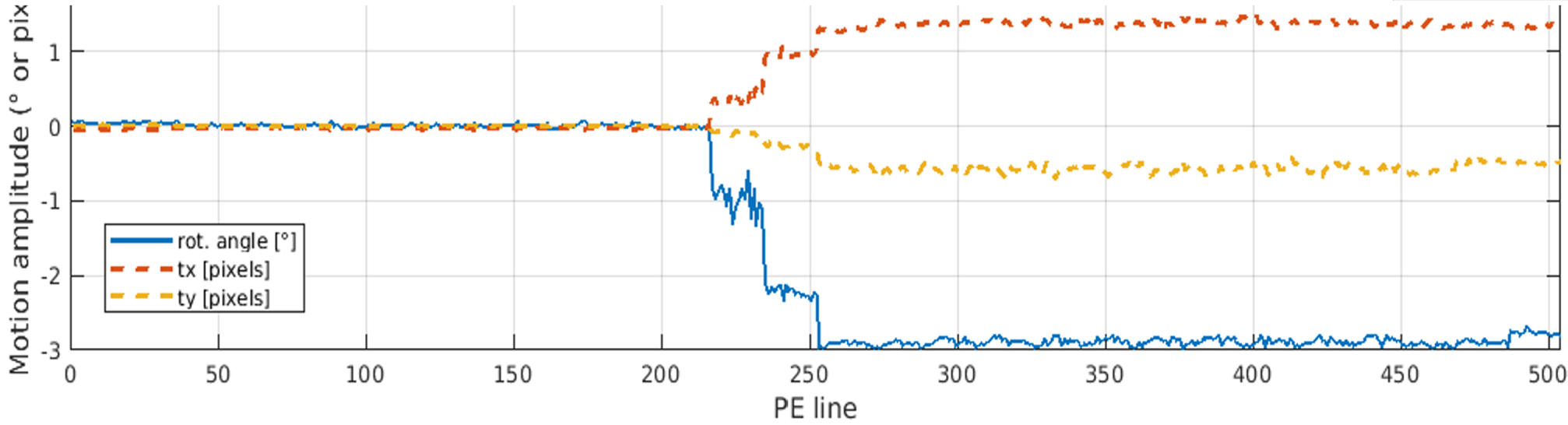}
        \caption{ MoSAIK-estimated motion trajectory of an in-vivo ChoCo motion-corrupted brain scan. }
        \label{fig:ph2}
    \end{subfigure}
	
	    \begin{subfigure}[t]{0.95\textwidth}
        \centering
        \includegraphics[width=\linewidth]{  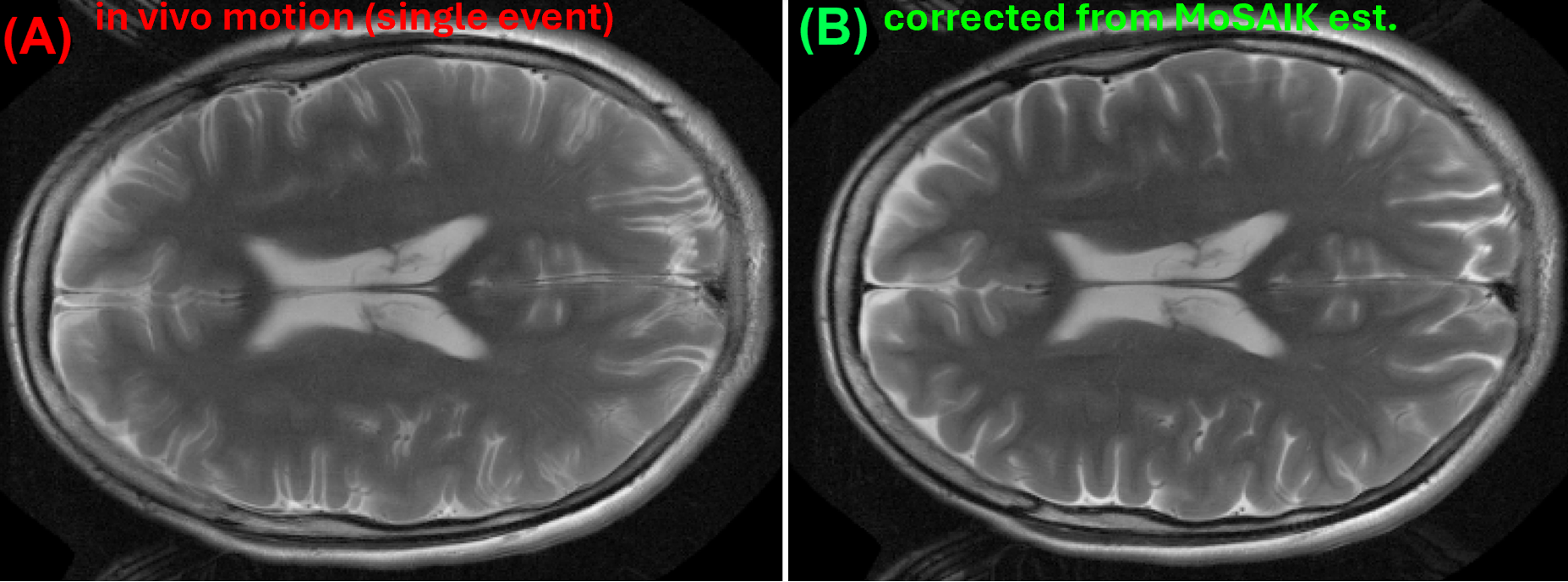}
        \caption{ Motion-corrected from the estimated trajectory in (b). }
        \label{fig:ph3}
    \end{subfigure}

    \caption{ PhRUITS: phantom and fruits trainingset and in vivo brain scans as a validation set.}
    \label{phruits}
\end{figure*}

The results also highlight the advantages of estimating motion in k-space, the native domain of MRI data acquisition. In reconstructed images, motion artefacts are spatially distributed, making it difficult to link them to specific acquisition inconsistencies. In k-space, these inconsistencies can instead be associated with the phase-encoding lines acquired at each motion state, which form characteristic echo-train ''combs''. This direct relationship with the acquisition process makes k-space a natural domain for learning motion-related signatures and informing motion correction. Learning these signatures may also reduce dependence on image contrast and anatomical content. Preliminary support for this comes from an experiment in which MoSAIK was trained exclusively on non-brain phantom data, yet successfully reduced motion artefacts in in vivo brain images (\autoref{phruits}). This finding suggests that the network learns transferable features of motion-induced k-space inconsistencies, rather than relying solely on anatomy-specific features.

\begin{figure*}[!t]
    \centering

    \begin{subfigure}[t]{0.95\textwidth}
        \centering
        \includegraphics[width=\linewidth]{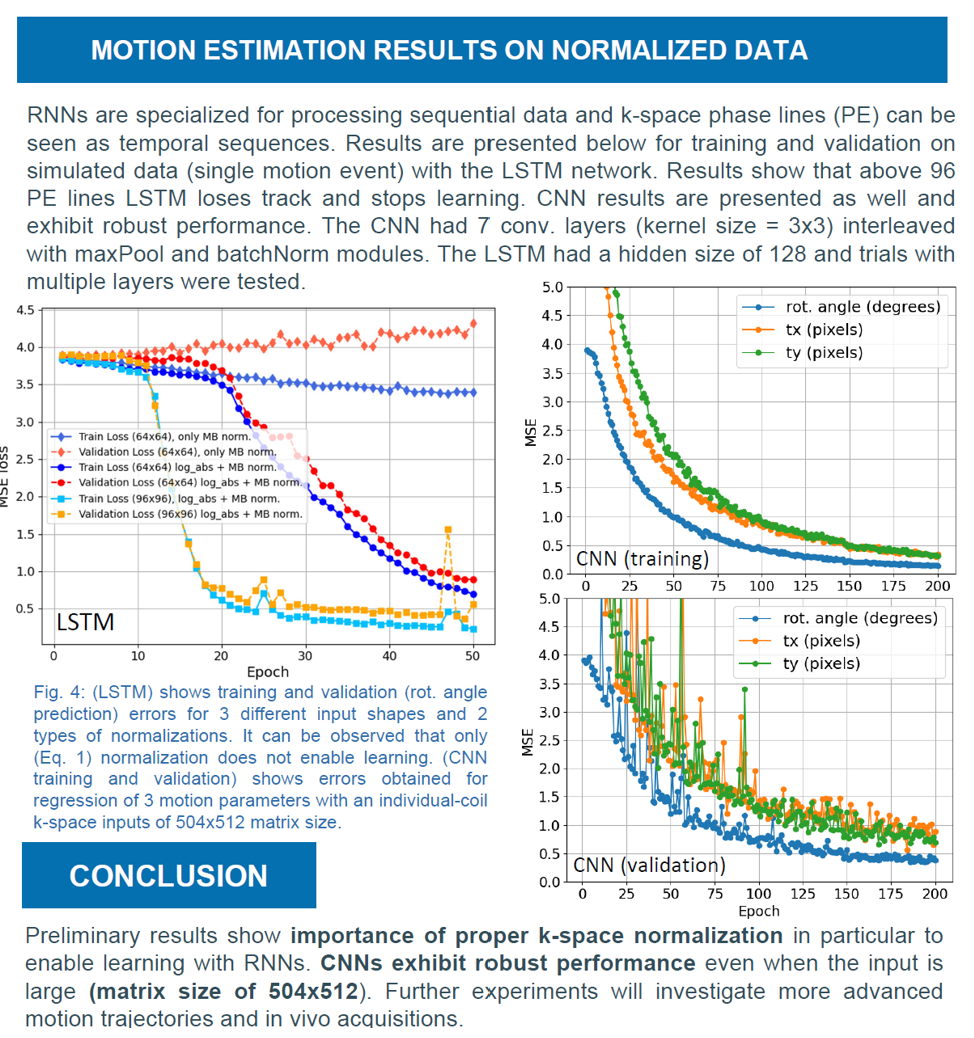}
    \end{subfigure}
    \caption{Preliminary results obtained with classical CNN and LSTM models were presented at the Alpine Brain Imaging Meeting (ABIM) 2024. These models were able to learn and generalize in the presence of \textbf{\textit{a single}} random motion event, which corresponds to the experimental setting shown in the plots presented here. However, performance was significantly reduced when multiple motion events were introduced, suggesting that their limited ability to capture long-range dependencies across k-space lines prevents them from accurately modeling more complex motion trajectories.}
    \label{cnn_lstm}
\end{figure*}

Preliminary experiments with CNN-based and RNN/LSTM-based models also informed the final architecture (\autoref{cnn_lstm}). Fully convolutional networks trained on simulated data with a single motion event placed at the middle of the scan, and using matrix sizes of (64 $\times$ 64), (128 $\times$ 128), and (256 $\times$ 256), were able to estimate $\theta)$,$t_x$, and $t_y$ for this simplified setting. However, CNN-only architectures performed poorly when multiple motion events were introduced. This is likely because local convolutional filtering alone is not sufficient to capture global relationships across distant k-space lines and echo-train blocks. RNN/LSTM models were also limited by the need to model long-range dependencies across the acquisition. In contrast, the transformer backbone can use self-attention to relate distant parts of the acquisition, which is better suited to multi-event motion trajectories. The ablation analysis further supported the contribution of the convolutional embedding module to generalization. Removing this module resulted in severe overfitting: the model fitted the training data almost perfectly but failed to generalize to unseen data. This suggests that extracting local k-space features before integrating information across the acquisition through self-attention provides an important inductive bias for motion estimation.

Several limitations remain. First, in vivo validation lacks ground-truth motion trajectories, so quantitative evaluation relies on indirect image-quality metrics and qualitative assessment. For each ChoCo acquisition, a motion-free scan was acquired before the motion-corrupted anatomical scans and could therefore serve as a pseudo-reference (as shown in \autoref{pseudoref}). However, this pseudo-reference is not guaranteed to match the corrupted acquisition exactly, since slice positioning and anatomical content may differ slightly between scans. Such differences can bias EFC-based comparisons and are even more problematic for full-reference metrics. For this reason, the use of these pseudo-references for quantitative evaluation was not included in the present analysis and remains under investigation.

Other referenceless image quality metrics were investigated, including Tenengrad (edge strength) and gradient entropy (which is essnetially EFC applied to  image gradients). Tenengrad provides information about image sharpness, whereas gradient entropy captures changes related to ghosting and blurring in a somewhat similar manner to EFC. However, Tenengrad generally increased after CG-SENSE reconstruction but usually decreased after TV reconstructions, probably because the latter produces smoother images. Gradient entropy also did not appear to reflect motion-correction quality accurately: it systematically indicated an improvement for TV reconstructions, resulting in an unrealistically favorable assessment that was not consistent with the qualitatively observed degree of motion correction. Therefore, we decided not to include Tenengrad and gradient entropy and retained EFC as our reference-free metric, as it showed the most consistent behavior across reconstruction methods evaluated in our experiments.

Second, the current validation was performed on a limited number of volunteers and prescribed motion conditions. Third, the model currently assumes 2D in-plane rigid motion, whereas real head motion may include through-plane components (although they seem to be far less prone to cause significant artefacts) and more complex temporal patterns. Finally, additional variability in acquisition protocols, including different turbo factors, contrasts, scanners, and coil configurations, should be included in future training and validation datasets.

Future work will therefore focus on increasing the diversity of simulated and in vivo training data, including stronger and more complex motion trajectories with multiple events, while preserving clinically realistic motion ranges. Further validation across scanners, pulse-sequence settings, and patient populations will also be investigated. Ultimately, integrating MoSAIK into a clinical reconstruction pipeline will require robust generalization across acquisition parameters and reliable handling of free-motion cases.

\section{Conclusion}
MoSAIK enables reference-free retrospective motion correction in 2D TSE/FSE MRI by estimating rigid in-plane motion directly from k-space followed by model-based reconstruction, thus avoiding deep hallucinations by design. By working in k-space, the method exploits localized motion-induced inconsistencies while reducing dependence on image contrast and anatomical content. Simulated and in vivo results showed consistent artefact reduction, supporting the potential of this approach for clinically relevant motion correction.

\section*{Acknowledgment}
We acknowledge the facilities and expertise of the CIBM Center for Biomedical Imaging, a Swiss research center of excellence founded and supported by CHUV, UNIL, EPFL, UNIGE, and HUG. The authors thank the volunteers and the HUG staff involved in the in-vivo MRI acquisitions. Author O.D. and author S.C. acknowledge that a patent application covering the core framework of this technology has been filed at the European Patent Office.

\section*{Declarations}

\subsection*{Funding}
This work was supported by the INNOGAP Proof-of-Principle Fund, University of Geneva and Geneva University Hospitals and by the Startup grant by the Department of Radiology and Medical Informatics, Faculty of Medicine, University of Geneva (DRIM).

\subsection*{Competing interests}
The authors declare that they have no conflicts of interest.

\subsection*{Data availability}
Volunteer and phantom data can be made available upon reasonable request.




\bibliography{sn-bibliography}

@article{andre2015toward,
  title={Toward quantifying the prevalence, severity, and cost associated with patient motion during clinical {MR} examinations},
  author={Andre, Jalal B and Bresnahan, Brian W and Mossa-Basha, Mahmud and Hoff, Michael N and Smith, C Patrick and Anzai, Yoshimi and Cohen, Wendy A},
  journal={Journal of the American College of Radiology},
  volume={12},
  number={7},
  pages={689--695},
  year={2015},
  publisher={Elsevier}
}

@article{slipsager2020quantifying,
  title={Quantifying the financial savings of motion correction in brain {MRI}: {A} model-based estimate of the costs arising from patient head motion and potential savings from implementation of motion correction},
  author={Slipsager, Jakob M and Glimberg, Stefan L and S{\o}gaard, Jes and Paulsen, Rasmus R and Johannesen, Helle H and Martens, Pernille C and Seth, Alka and Marner, Lisbeth and Henriksen, Otto M and Olesen, Oline V and others},
  journal={Journal of Magnetic Resonance Imaging},
  volume={52},
  number={3},
  pages={731--738},
  year={2020},
  publisher={Wiley Online Library}
}

@article{hennig1986rare,
  title={{RARE} imaging: {A} fast imaging method for clinical {MR}},
  author={Hennig, J{\"u}rgen and Nauerth, A and Friedburg, HRARE},
  journal={Magnetic resonance in medicine},
  volume={3},
  number={6},
  pages={823--833},
  year={1986},
  publisher={Wiley Online Library}
}

@article{pipe1999motion,
  title={Motion correction with {PROPELLER MRI}: {Application} to head motion and free-breathing cardiac imaging},
  author={Pipe, James G},
  journal={Magnetic Resonance in Medicine: An Official Journal of the International Society for Magnetic Resonance in Medicine},
  volume={42},
  number={5},
  pages={963--969},
  year={1999},
  publisher={Wiley Online Library}
}

@article{ehman1989adaptive,
  title={Adaptive technique for high-definition {MR} imaging of moving structures.},
  author={Ehman, Richard L and Felmlee, Joel P},
  journal={Radiology},
  volume={173},
  number={1},
  pages={255--263},
  year={1989}
}

@article{johnson2016retrospective,
  title={Retrospective {3D} motion correction using spherical navigator echoes},
  author={Johnson, Patricia M and Liu, Junmin and Wade, Trevor and Tavallaei, Mohammad Ali and Drangova, Maria},
  journal={Magnetic resonance imaging},
  volume={34},
  number={9},
  pages={1274--1282},
  year={2016},
  publisher={Elsevier}
}

@article{tremblay2005retrospective,
  title={Retrospective coregistration of functional magnetic resonance imaging data using external monitoring},
  author={Tremblay, Marleine and Tam, Fred and Graham, Simon J},
  journal={Magnetic Resonance in Medicine: An Official Journal of the International Society for Magnetic Resonance in Medicine},
  volume={53},
  number={1},
  pages={141--149},
  year={2005},
  publisher={Wiley Online Library}
}

@article{afacan2020retrospective,
  title={Retrospective correction of head motion using measurements from an electromagnetic tracker},
  author={Afacan, Onur and Wallace, Tess E and Warfield, Simon K},
  journal={Magnetic resonance in medicine},
  volume={83},
  number={2},
  pages={427--437},
  year={2020},
  publisher={Wiley Online Library}
}

@article{atkinson1997automatic,
  title={Automatic correction of motion artifacts in magnetic resonance images using an entropy focus criterion},
  author={Atkinson, David and Hill, Derek LG and Stoyle, Peter NR and Summers, Paul E and Keevil, Steven F},
  journal={IEEE Transactions on Medical imaging},
  volume={16},
  number={6},
  pages={903--910},
  year={1997},
  publisher={IEEE}
}

@article{atkinson1999automatic,
  title={Automatic compensation of motion artifacts in {MRI}},
  author={Atkinson, David and Hill, Derek LG and Stoyle, Peter NR and Summers, Paul E and Clare, Stuart and Bowtell, Richard and Keevil, Stephen F},
  journal={Magnetic Resonance in Medicine: An Official Journal of the International Society for Magnetic Resonance in Medicine},
  volume={41},
  number={1},
  pages={163--170},
  year={1999},
  publisher={Wiley Online Library}
}

@article{kustner2019retrospective,
  title={Retrospective correction of motion-affected {MR} images using deep learning frameworks},
  author={K{\"u}stner, Thomas and Armanious, Karim and Yang, Jiahuan and Yang, Bin and Schick, Fritz and Gatidis, Sergios},
  journal={Magnetic resonance in medicine},
  volume={82},
  number={4},
  pages={1527--1540},
  year={2019},
  publisher={Wiley Online Library}
}

@article{pawar2022suppressing,
  title={Suppressing motion artefacts in {MRI} using an {Inception-ResNet} network with motion simulation augmentation},
  author={Pawar, Kamlesh and Chen, Zhaolin and Shah, N Jon and Egan, Gary F},
  journal={NMR in Biomedicine},
  volume={35},
  number={4},
  pages={e4225},
  year={2022},
  publisher={Wiley Online Library}
}

@article{kang2024motion,
  title={Motion artifact reduction using {U-net} model with three-dimensional simulation-based datasets for brain magnetic resonance images},
  author={Kang, Seong-Hyeon and Lee, Youngjin},
  journal={Bioengineering},
  volume={11},
  number={3},
  pages={227},
  year={2024},
  publisher={MDPI}
}

@article{armanious2020medgan,
  title={{MedGAN}: {Medical} image translation using {GANs}},
  author={Armanious, Karim and Jiang, Chenming and Fischer, Marc and K{\"u}stner, Thomas and Hepp, Tobias and Nikolaou, Konstantin and Gatidis, Sergios and Yang, Bin},
  journal={Computerized medical imaging and graphics},
  volume={79},
  pages={101684},
  year={2020},
  publisher={Elsevier}
}

@article{johnson2019conditional,
  title={Conditional generative adversarial network for {3D} rigid-body motion correction in {MRI}},
  author={Johnson, Patricia M and Drangova, Maria},
  journal={Magnetic resonance in medicine},
  volume={82},
  number={3},
  pages={901--910},
  year={2019},
  publisher={Wiley Online Library}
}

@article{haskell2019network,
  title={Network accelerated motion estimation and reduction ({NAMER}): {Convolutional} neural network guided retrospective motion correction using a separable motion model},
  author={Haskell, Melissa W and Cauley, Stephen F and Bilgic, Berkin and Hossbach, Julian and Splitthoff, Daniel N and Pfeuffer, Josef and Setsompop, Kawin and Wald, Lawrence L},
  journal={Magnetic resonance in medicine},
  volume={82},
  number={4},
  pages={1452--1461},
  year={2019},
  publisher={Wiley Online Library}
}

@article{dabrowski2024sismik,
  title={{SISMIK} for brain {MRI}: {Deep}-learning-based motion estimation and model-based motion correction in k-space},
  author={Dabrowski, Oscar and Falcone, Jean-Luc and Klauser, Antoine and Songeon, Julien and Kocher, Michel and Chopard, Bastien and Lazeyras, Fran{\c{c}}ois and Courvoisier, S{\'e}bastien},
  journal={IEEE Transactions on Medical Imaging},
  year={2024},
  publisher={IEEE}
}

@article{hossbach2023deep,
  title={Deep learning-based motion quantification from k-space for fast model-based magnetic resonance imaging motion correction},
  author={Hossbach, Julian and Splitthoff, Daniel Nicolas and Cauley, Stephen and Clifford, Bryan and Polak, Daniel and Lo, Wei-Ching and Meyer, Heiko and Maier, Andreas},
  journal={Medical physics},
  volume={50},
  number={4},
  pages={2148--2161},
  year={2023},
  publisher={Wiley Online Library}
}

@inproceedings{eichhorn2023physics,
  title={Physics-aware motion simulation for {T2*}-weighted brain {MRI}},
  author={Eichhorn, Hannah and Hammernik, Kerstin and Spieker, Veronika and Epp, Samira M and Rueckert, Daniel and Preibisch, Christine and Schnabel, Julia A},
  booktitle={International Workshop on Simulation and Synthesis in Medical Imaging},
  pages={42--52},
  year={2023},
  organization={Springer}
}

@inproceedings{chen2021retrospective,
  title={Retrospective motion correction for Fast Spin Echo based on conditional {GAN} with entropy loss},
  author={Chen, Yalei and Li, Pingan and Liu, Kewen and Bao, Qingjia and Li, Zhao and Li, Xiaojun},
  booktitle={Journal of Physics: Conference Series},
  volume={1848},
  number={1},
  pages={012118},
  year={2021},
  organization={IOP Publishing}
}

@inproceedings{zhang2024reducing,
  title={Reducing motion artifacts in brain {MRI} using vision transformers and self-supervised learning},
  author={Zhang, Lei and Wang, Xiaoke and Herskovits, Edward H and Melhem, Elias R and Chang, Linda and Wang, Ze and Ernst, Thomas},
  booktitle={2024 IEEE International Conference on Image Processing (ICIP)},
  pages={3165--3171},
  year={2024},
  organization={IEEE}
}

@article{maclaren2013prospective,
  title={Prospective motion correction in brain imaging: {A} review},
  author={Maclaren, Julian and Herbst, Michael and Speck, Oliver and Zaitsev, Maxim},
  journal={Magnetic resonance in medicine},
  volume={69},
  number={3},
  pages={621--636},
  year={2013},
  publisher={Wiley Online Library}
}

@article{loktyushin2013blind,
  title={Blind retrospective motion correction of {MR} images},
  author={Loktyushin, Alexander and Nickisch, Hannes and Pohmann, Rolf and Sch{\"o}lkopf, Bernhard},
  journal={Magnetic resonance in medicine},
  volume={70},
  number={6},
  pages={1608--1618},
  year={2013},
  publisher={Wiley Online Library}
}

@article{loktyushin2015blind,
  title={Blind multirigid retrospective motion correction of {MR} images},
  author={Loktyushin, Alexander and Nickisch, Hannes and Pohmann, Rolf and Sch{\"o}lkopf, Bernhard},
  journal={Magnetic resonance in medicine},
  volume={73},
  number={4},
  pages={1457--1468},
  year={2015},
  publisher={Wiley Online Library}
}

@inproceedings{singh2024data,
  title={Data consistent deep rigid {MRI} motion correction},
  author={Singh, Nalini M and Dey, Neel and Hoffmann, Malte and Fischl, Bruce and Adalsteinsson, Elfar and Frost, Robert and Dalca, Adrian V and Golland, Polina},
  booktitle={Medical imaging with deep learning},
  pages={368--381},
  year={2024},
  organization={PMLR}
}

@article{lee2024unsupervised,
  title={Unsupervised motion artifact correction of turbo spin-echo {MRI} using deep image prior},
  author={Lee, Jongyeon and Seo, Hyunseok and Lee, Wonil and Park, HyunWook},
  journal={Magnetic Resonance in Medicine},
  volume={92},
  number={1},
  pages={28--42},
  year={2024},
  publisher={Wiley Online Library}
}

@article{dmitry2020deep,
  title={Deep image prior},
  author={Dmitry, Ulyanov and Vedaldi, Andrea and Victor, Lempitsky},
  journal={International Journal of Computer Vision},
  volume={128},
  number={7},
  pages={1867--1888},
  year={2020},
  publisher={Springer Nature BV}
}

@article{safari2025physics,
  title={A Physics-Informed Deep Learning Model for {MRI} Brain Motion Correction},
  author={Safari, Mojtaba and Wang, Shansong and Eidex, Zach and Qiu, Richard and Chang, Chih-Wei and Yu, David S and Yang, Xiaofeng},
  journal={ArXiv},
  pages={arXiv--2502},
  year={2025}
}

@article{cui2023motion,
  title={Motion artifact reduction for magnetic resonance imaging with deep learning and k-space analysis},
  author={Cui, Long and Song, Yang and Wang, Yida and Wang, Rui and Wu, Dongmei and Xie, Haibin and Li, Jianqi and Yang, Guang},
  journal={PloS one},
  volume={18},
  number={1},
  pages={e0278668},
  year={2023},
  publisher={Public Library of Science San Francisco, CA USA}
}

@article{bhadra2021hallucinations,
  title={On hallucinations in tomographic image reconstruction},
  author={Bhadra, Sayantan and Kelkar, Varun A and Brooks, Frank J and Anastasio, Mark A},
  journal={IEEE transactions on medical imaging},
  volume={40},
  number={11},
  pages={3249--3260},
  year={2021},
  publisher={IEEE}
}

@article{gottschling2025troublesome,
  title={The troublesome kernel: {On} hallucinations, no free lunches, and the accuracy-stability tradeoff in inverse problems},
  author={Gottschling, Nina M and Antun, Vegard and Hansen, Anders C and Adcock, Ben},
  journal={SIAM Review},
  volume={67},
  number={1},
  pages={73--104},
  year={2025},
  publisher={SIAM}
}

@article{vaswani2017attention,
  title={Attention is all you need},
  author={Vaswani, Ashish and Shazeer, Noam and Parmar, Niki and Uszkoreit, Jakob and Jones, Llion and Gomez, Aidan N and Kaiser, {\L}ukasz and Polosukhin, Illia},
  journal={Advances in neural information processing systems},
  volume={30},
  year={2017}
}

@inproceedings{DosovitskiyB0WZ21,
  author       = {Alexey Dosovitskiy and
                  Lucas Beyer and
                  Alexander Kolesnikov and
                  Dirk Weissenborn and
                  Xiaohua Zhai and
                  Thomas Unterthiner and
                  Mostafa Dehghani and
                  Matthias Minderer and
                  Georg Heigold and
                  Sylvain Gelly and
                  Jakob Uszkoreit and
                  Neil Houlsby},
  title        = {An Image is Worth 16x16 Words: {Transformers} for Image Recognition
                  at Scale},
  booktitle    = {9th International Conference on Learning Representations, {ICLR} 2021,
                  Virtual Event, Austria, May 3-7, 2021},
  publisher    = {OpenReview.net},
  year         = {2021},
  url          = {https://openreview.net/forum?id=YicbFdNTTy},
  bibsource    = {dblp computer science bibliography, https://dblp.org}
}

@Article{MaskedAutoencoders2021,
  author  = {Kaiming He and Xinlei Chen and Saining Xie and Yanghao Li and Piotr Doll{\'a}r and Ross Girshick},
  journal = {arXiv:2111.06377},
  title   = {Masked Autoencoders Are Scalable Vision Learners},
  year    = {2021},
}

@article{bhuiyan2023feasibility,
  title={Feasibility of tracking involuntary head movement for {MRI} using a coil as a magnetic dipole in a time-varying gradient},
  author={Bhuiyan, EH and Chowdhury, MEH and Glover, PM},
  journal={Magnetic Resonance Imaging},
  volume={101},
  pages={76--89},
  year={2023},
  publisher={Elsevier}
}

@article{bhuiyan2024tracking,
  title={Tracking head movement inside an {MR} scanner using electromagnetic coils},
  author={Bhuiyan, Enamul Hoque and Chowdhury, Muhammad Enamul Hoque and Glover, Paul M},
  journal={Heliyon},
  volume={10},
  number={13},
  year={2024},
  publisher={Elsevier}
}

@article{batchelor2005matrix,
  title={Matrix description of general motion correction applied to multishot images},
  author={Batchelor, PG and Atkinson, D and Irarrazaval, P and Hill, DLG and Hajnal, J and Larkman, D},
  journal={Magnetic Resonance in Medicine: An Official Journal of the International Society for Magnetic Resonance in Medicine},
  volume={54},
  number={5},
  pages={1273--1280},
  year={2005},
  publisher={Wiley Online Library}
}

@inproceedings{xiong2020layer,
  title={On layer normalization in the transformer architecture},
  author={Xiong, Ruibin and Yang, Yunchang and He, Di and Zheng, Kai and Zheng, Shuxin and Xing, Chen and Zhang, Huishuai and Lan, Yanyan and Wang, Liwei and Liu, Tieyan},
  booktitle={International conference on machine learning},
  pages={10524--10533},
  year={2020},
  organization={PMLR}
}

@article{tachella2025deepinverse,
    title = {DeepInverse: {A} Python package for solving imaging inverse problems with deep learning},
    journal = {Journal of Open Source Software},
    doi = {10.21105/joss.08923},
    url = {https://doi.org/10.21105/joss.08923},
    year = {2025},
    publisher = {The Open Journal},
    volume = {10},
    number = {115},
    pages = {8923},
    author = {Tachella, Julián and Terris, Matthieu and Hurault, Samuel and Wang, Andrew and Davy, Leo and Scanvic, Jérémy and Sechaud, Victor and Vo, Romain and Moreau, Thomas and Davies, Thomas and Chen, Dongdong and Laurent, Nils and Monroy, Brayan and Dong, Jonathan and Hu, Zhiyuan and Nguyen, Minh-Hai and Sarron, Florian and Weiss, Pierre and Escande, Paul and Massias, Mathurin and Modrzyk, Thibaut and Levac, Brett and Liaudat, Tobías I. and Song, Maxime and Hertrich, Johannes and Neumayer, Sebastian and Schramm, Georg},
}

@article{dabrowski2022choreography,
  title={Choreography Controlled ({ChoCo}) brain {MRI} artifact generation for labeled motion-corrupted datasets},
  author={Dabrowski, Oscar and Courvoisier, S{\'e}bastien and Falcone, Jean-Luc and Klauser, Antoine and Songeon, Julien and Kocher, Michel and Chopard, Bastien and Lazeyras, Fran{\c{c}}ois},
  journal={Physica Medica},
  volume={102},
  pages={79--87},
  year={2022},
  publisher={Elsevier}
}

@inproceedings{baritaux2010priori,
  title={A priori guided reconstruction for fdot using mixed norms},
  author={Baritaux, Jean-Charles and Unser, Michael},
  booktitle={2010 IEEE International Symposium on Biomedical Imaging: From Nano to Macro},
  pages={692--695},
  year={2010},
  organization={IEEE}
}

@conference{muckley20tah,
  author = {M. J. Muckley and R. Stern and T. Murrell and F. Knoll},
  title = {{TorchKbNufft}: A High-Level, Hardware-Agnostic Non-Uniform Fast {Fourier} Transform},
  booktitle = {ISMRM Workshop on Data Sampling \& Image Reconstruction},
  year = 2020,
  note = {Source code available at https://github.com/mmuckley/torchkbnufft},
}

\section{Appendix}
\section{In-vivo free motion}

\begin{figure*}[!ht]
    \centering

    \begin{subfigure}[t]{0.99\textwidth}
        \centering
        \includegraphics[width=\linewidth]{ 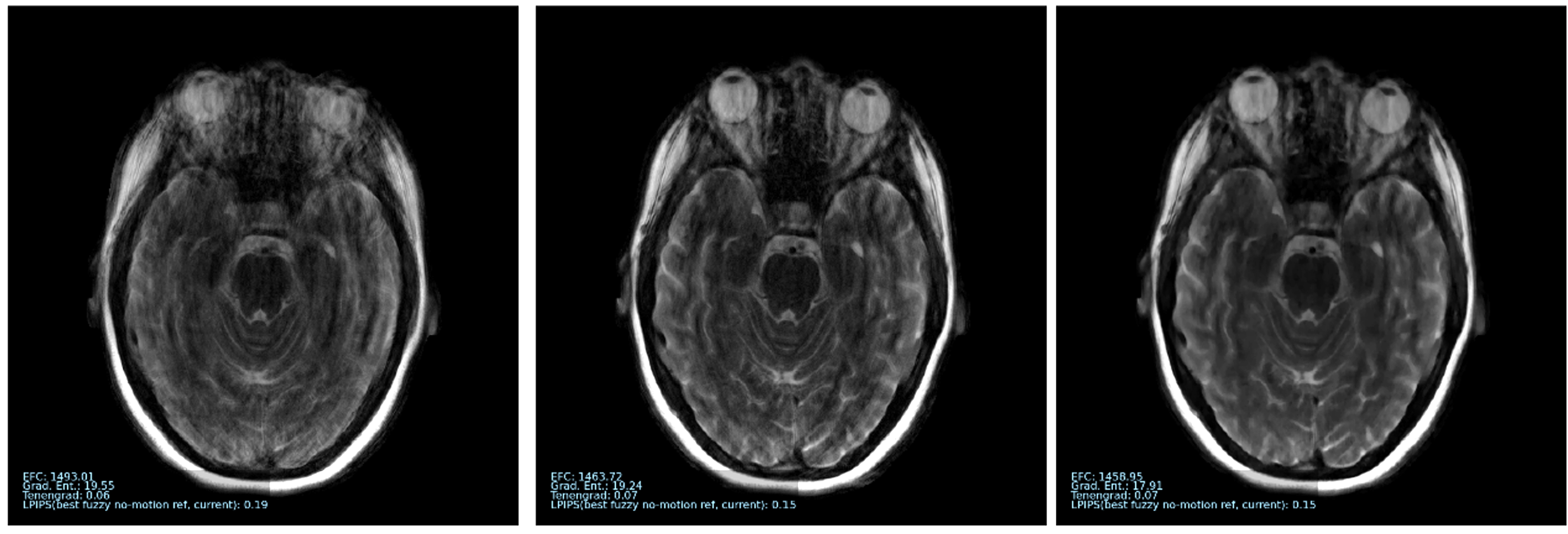 }
        \caption{ In-vivo free-motion acquisition: the subject was allowed to move freely during the scan. (Center) Reconstruction from estimated MoSAIK trajectory using Tikhonov-regularized CG-SENSE (c). (Right) Same as (Center) but with TV-regularization.  }
        \label{fig:invivofreemotion1_1}
    \end{subfigure}
    
      \vspace{2mm}

    \begin{subfigure}[t]{0.999\textwidth}
        \centering
        \includegraphics[width=\linewidth]{  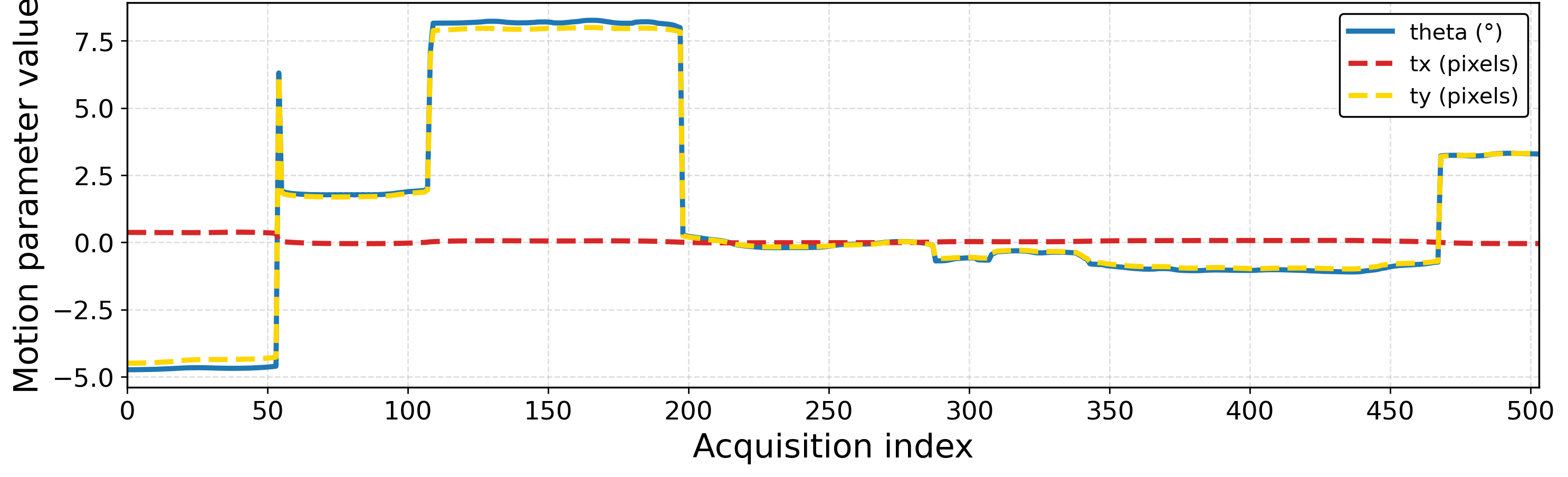}
        \caption{ MoSAIK-estimated motion trajectory (post-processed with median filtering and TV denoising). }
        \label{fig:invivofreemotion1_2}
    \end{subfigure}

    \caption{ In-vivo free-motion ChoCo acquisition. (a) Motion-corrupted image and MoSAIK motion-corrected reconstructions obtained with Tikhonov and TV regularization. (b) Post-processed MoSAIK-estimated motion trajectory during the free-motion scan.}
    \label{freeMotionSlice1}
\end{figure*}

\begin{figure*}[!t]
    \centering

    \begin{subfigure}[t]{0.99\textwidth}
        \centering
        \includegraphics[width=\linewidth]{ 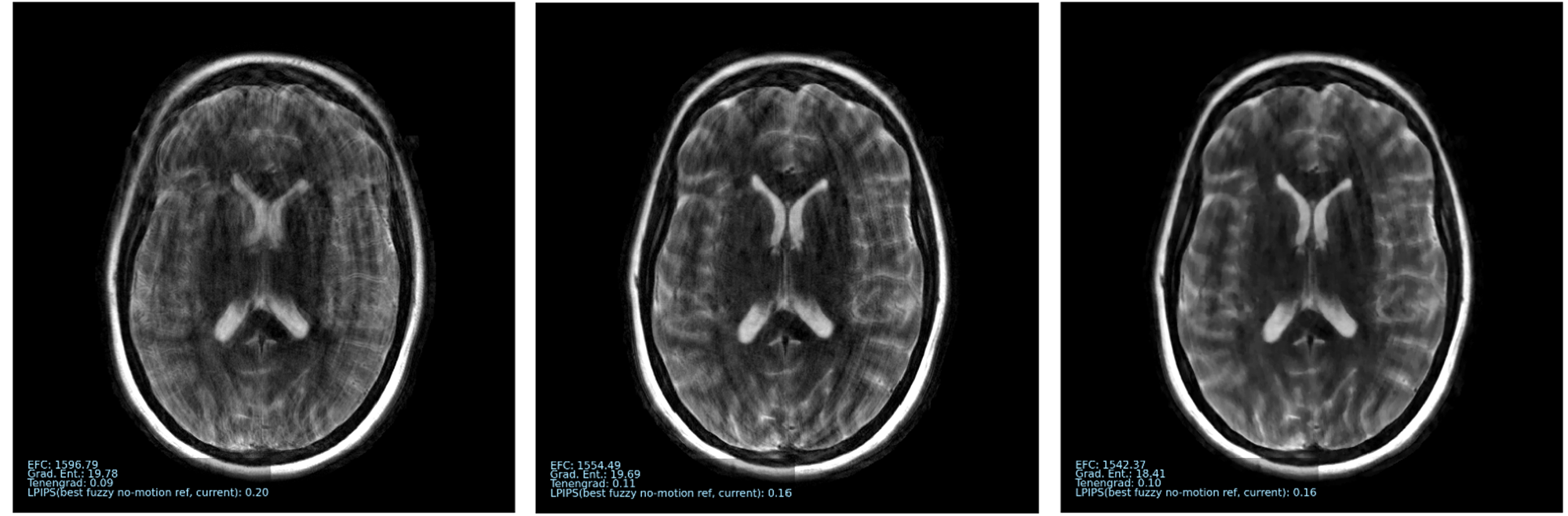 }
        \caption{In-vivo free-motion acquisition: the subject was allowed to move freely during the scan. Same subject as in \autoref{fig:invivofreemotion1_1}, shown on a different slice where the reconstruction is less successful. (Center)  Reconstruction from estimated MoSAIK trajectory using Tikhonov-regularized CG-SENSE (c). (Right) Same as (Center) but with TV-regularization.  }
        \label{fig:invivofreemotion2_1}
    \end{subfigure}
    
      \vspace{2mm}

    \begin{subfigure}[t]{0.999\textwidth}
        \centering
        \includegraphics[width=\linewidth]{  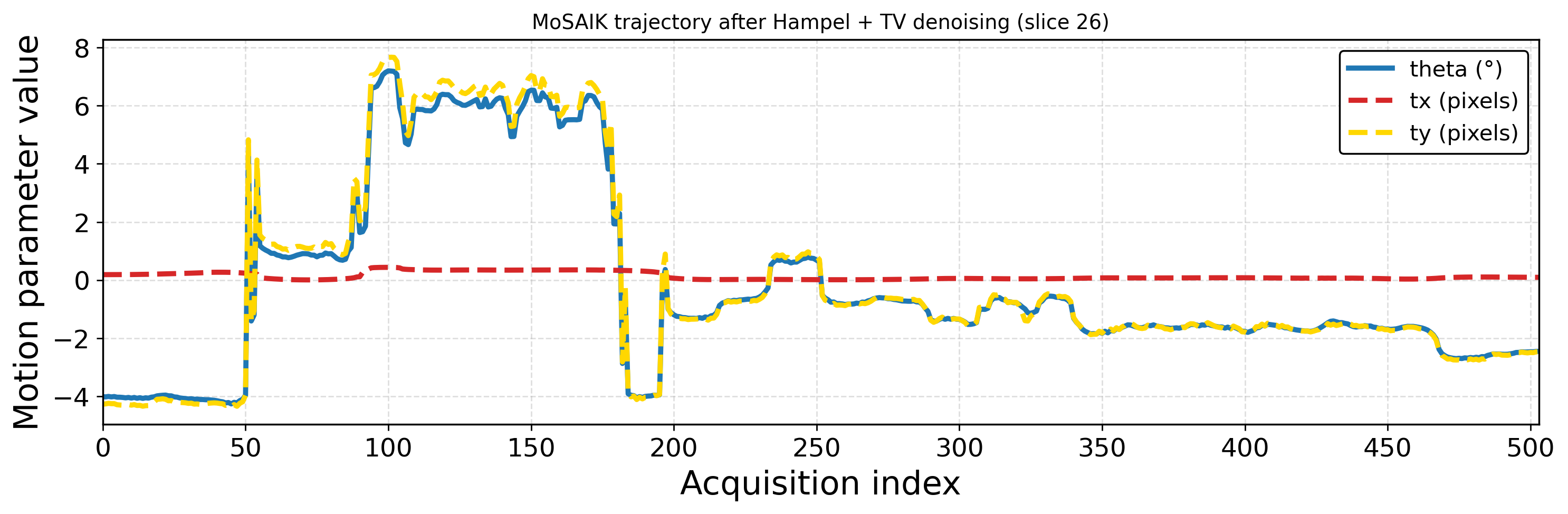}
        \caption{ MoSAIK-estimated motion trajectory (post-processed with median filtering and TV denoising). }
        \label{fig:invivofreemotion2_2}
    \end{subfigure}

    \caption{ In-vivo free-motion ChoCo acquisition. (a) Corrupted and MoSAIK-corrected images for a slice with residual artefacts. (b) Post-processed MoSAIK-estimated motion trajectory.}
    \label{freeMotionSlice2}
\end{figure*}


\begin{figure*}[!t]
    \centering
        \begin{subfigure}[t]{0.31\textwidth}
        \centering
        \includegraphics[width=\linewidth]{ 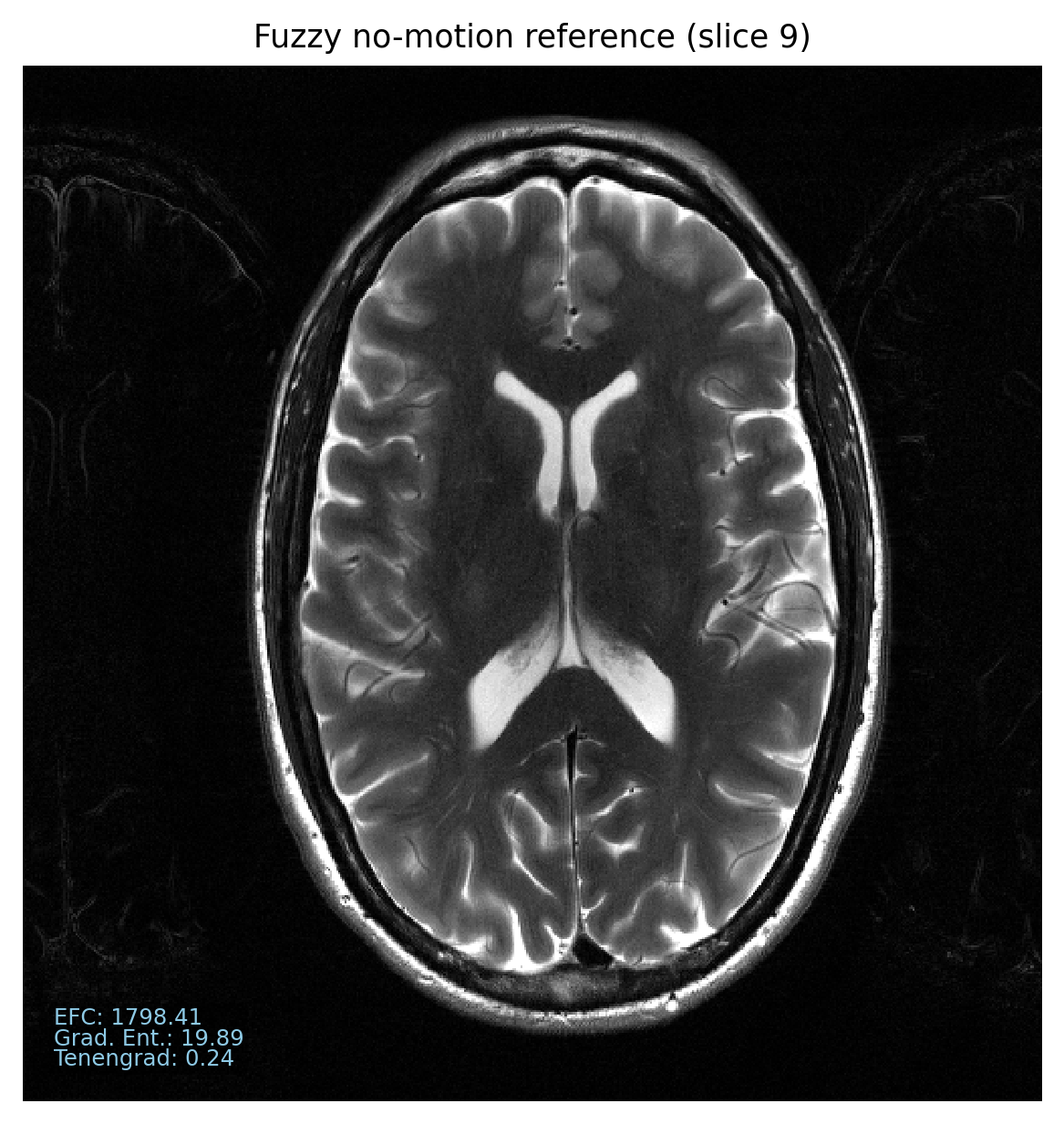 }
        \caption{ pseudo nomo ref GI .  }
        \label{fig:s3_nomo}
    \end{subfigure}
    \hspace{0.01\textwidth}
    \begin{subfigure}[t]{0.31\textwidth}
        \centering
        \includegraphics[width=\linewidth]{ 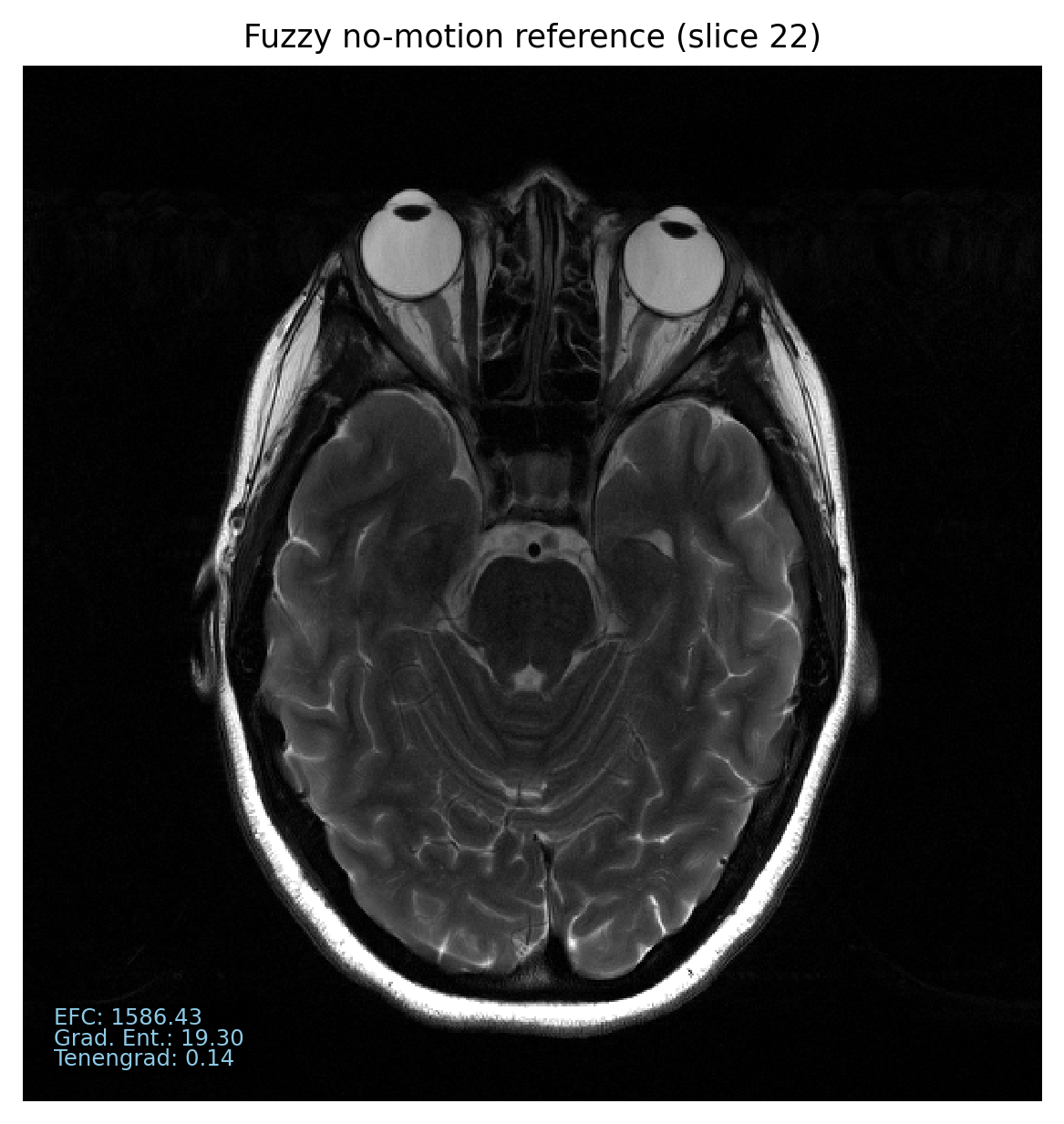 }
        \caption{ pseudo nomo ref 1.  }
        \label{fig:invivofreemotion1_ref}
    \end{subfigure}
    \hspace{0.01\textwidth}
    \begin{subfigure}[t]{0.31\textwidth}
        \centering
        \includegraphics[width=\linewidth]{  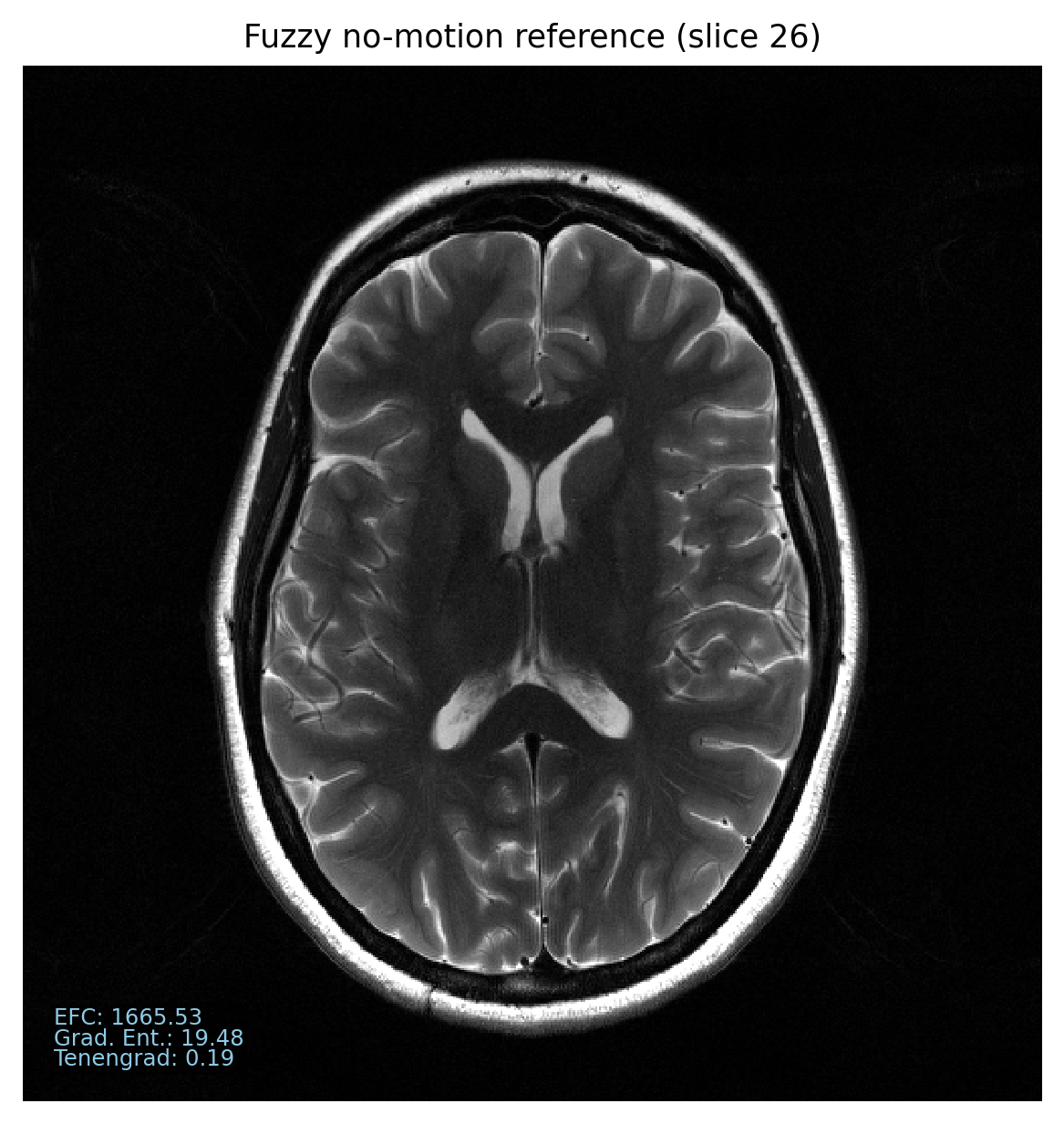}
        \caption{ pseudo nomo ref 2. }
        \label{fig:invivofreemotion2_ref}
    \end{subfigure}
    
    \caption{ Motion-free reference slices for: (a)  \autoref{invivoS3} and \autoref{simuS3ChocoIdeal}, (b) \autoref{freeMotionSlice1}, (c) \autoref{freeMotionSlice2}.}
    \label{pseudoref}
\end{figure*}

\begin{figure*}[!t]
    \centering

    \begin{subfigure}[t]{0.7\textwidth}
        \centering
        \includegraphics[width=\linewidth]{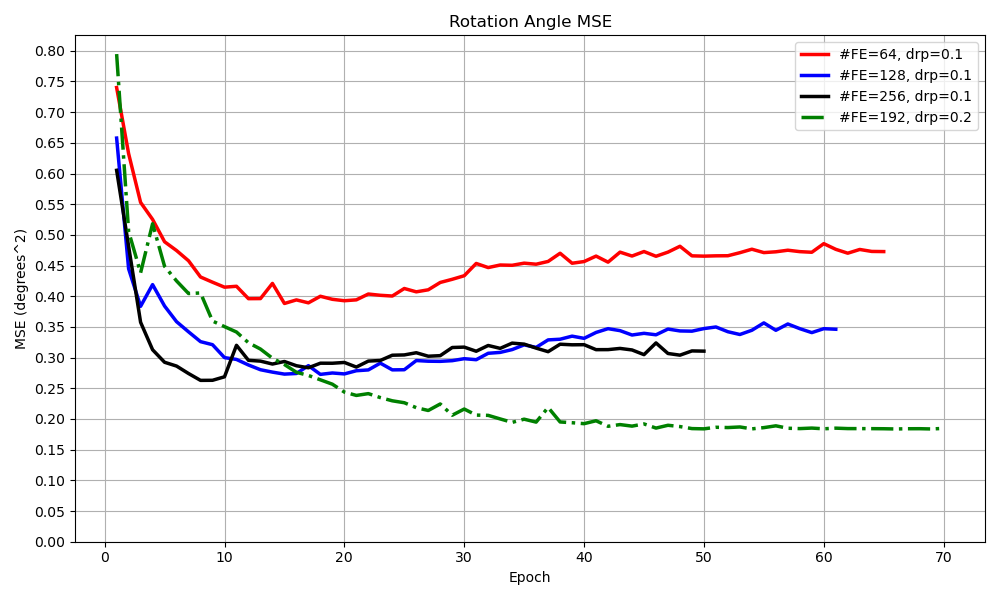}
    \end{subfigure}
    \vspace{2mm}
        \begin{subfigure}[t]{0.7\textwidth}
        \centering
        \includegraphics[width=\linewidth]{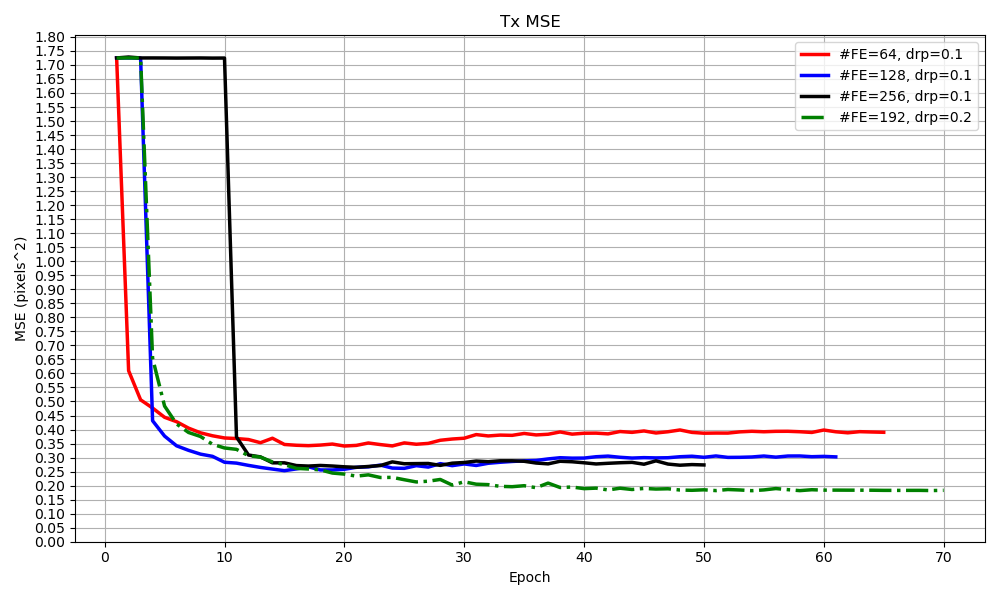}
    \end{subfigure}
    \vspace{2mm}
        \begin{subfigure}[t]{0.7\textwidth}
        \centering
        \includegraphics[width=\linewidth]{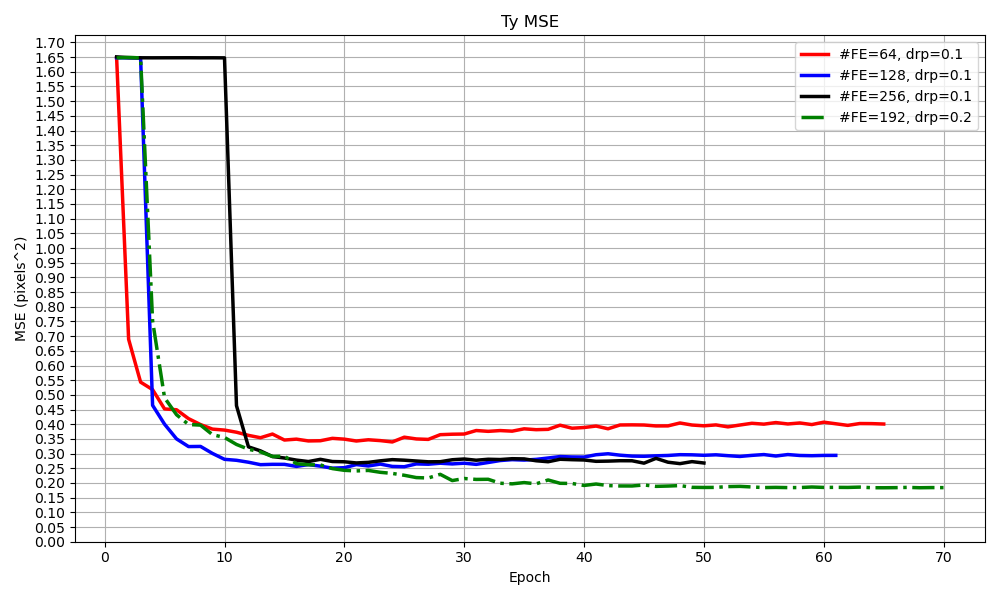}
    \end{subfigure}
    \caption{Validation-set mean squared error (MSE) plots between predicted and ground-truth motion parameters for MoSAIK models trained with different hyperparameter configurations. The lowest MSEs were achieved using a 192-line crop along the frequency-encoding direction and a dropout rate of 0.2.}
    \label{plots_val_MSE}
\end{figure*}

\begin{figure*}[!t]
    \centering

    \begin{subfigure}[t]{0.7\textwidth}
        \centering
        \includegraphics[width=\linewidth]{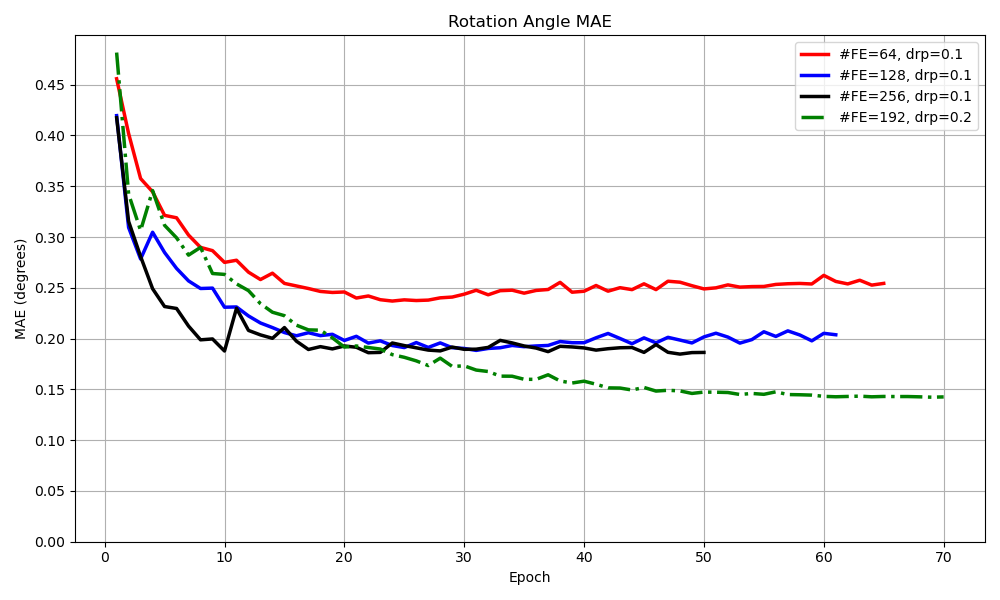}
    \end{subfigure}
    \vspace{2mm}
        \begin{subfigure}[t]{0.7\textwidth}
        \centering
        \includegraphics[width=\linewidth]{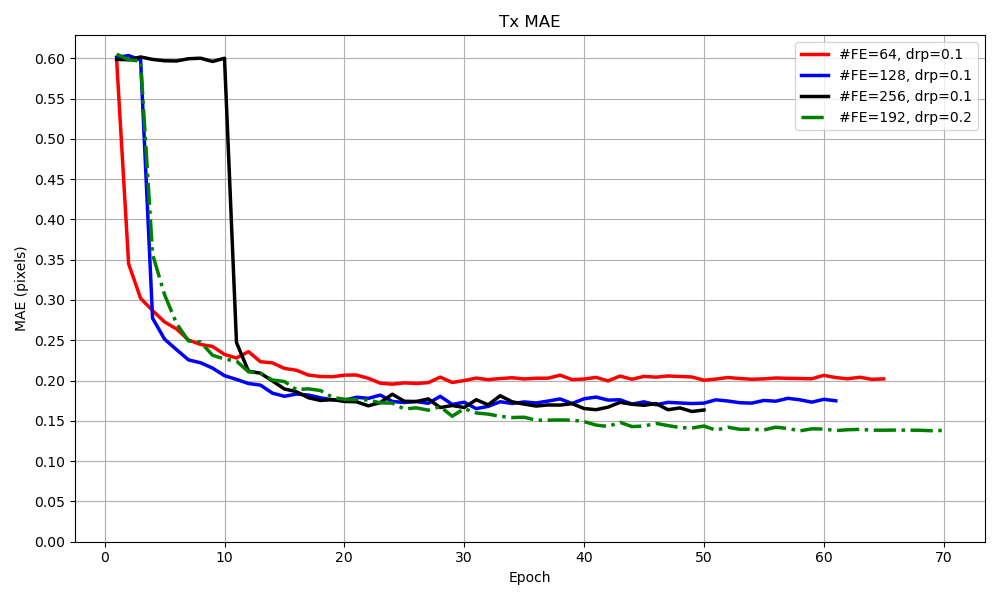}
    \end{subfigure}
    \vspace{2mm}
        \begin{subfigure}[t]{0.7\textwidth}
        \centering
        \includegraphics[width=\linewidth]{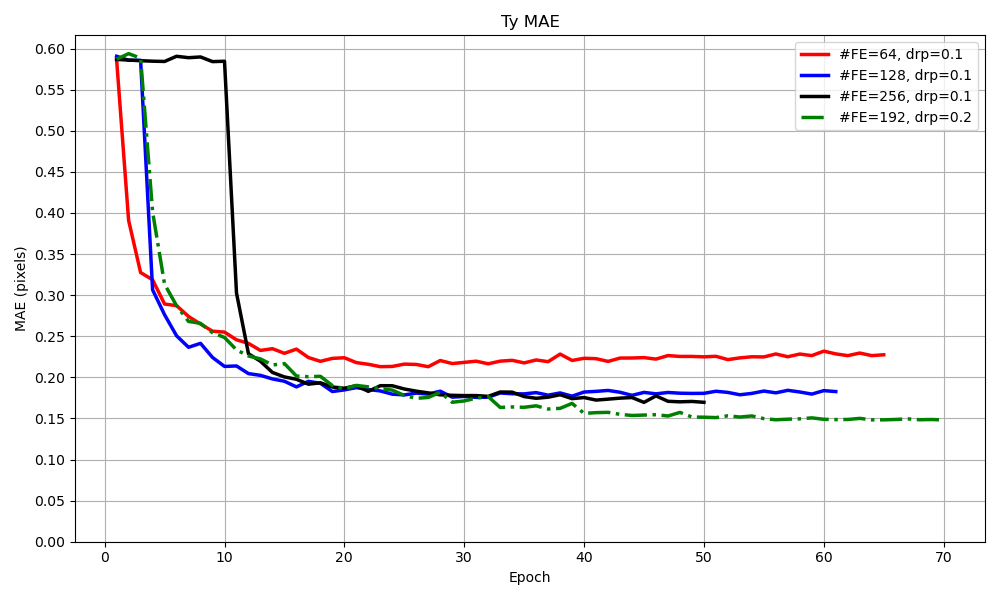}
    \end{subfigure}
    \caption{Validation-set mean absolute error (MAE) plots between predicted and ground-truth motion parameters for MoSAIK models trained with different hyperparameter configurations. The lowest MAEs were achieved using a 192-line crop along the frequency-encoding direction and a dropout rate of 0.2.}
    \label{plots_val_MAE}
\end{figure*}

\begin{figure*}[!t]
    \centering

    \begin{subfigure}[t]{0.7\textwidth}
        \centering
        \includegraphics[width=\linewidth]{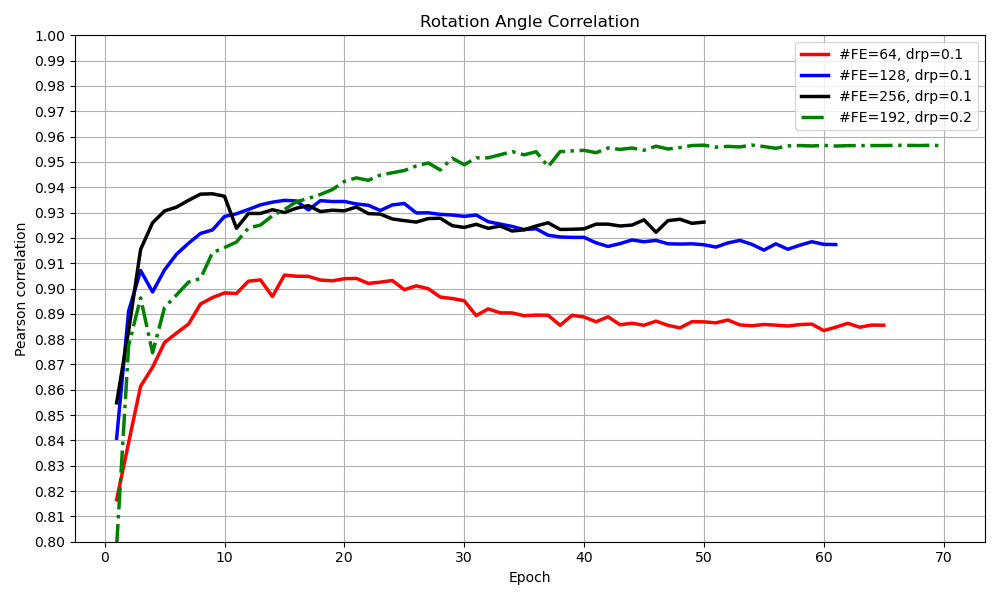}
    \end{subfigure}
    \vspace{2mm}
        \begin{subfigure}[t]{0.7\textwidth}
        \centering
        \includegraphics[width=\linewidth]{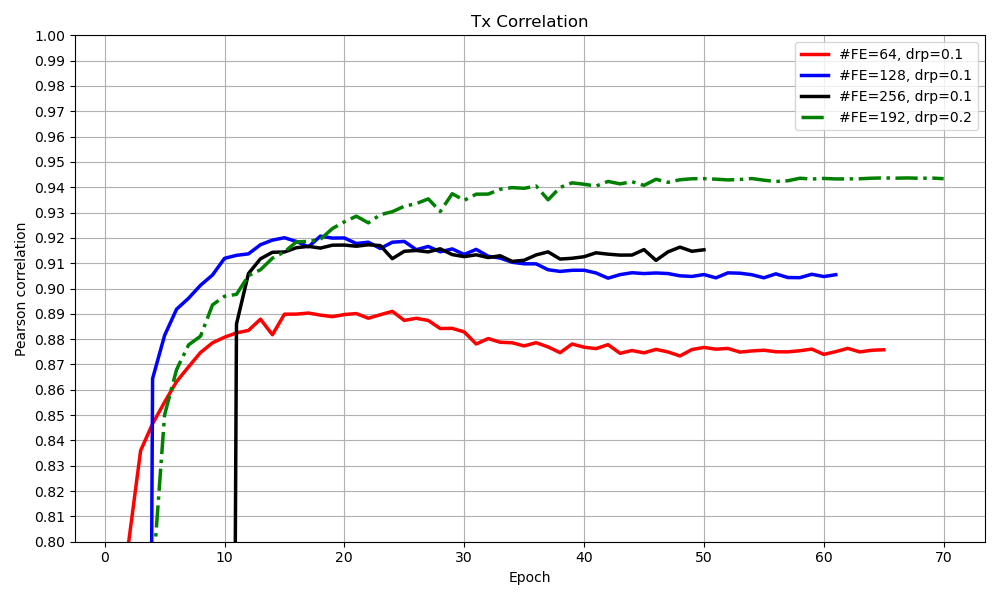}
    \end{subfigure}
    \vspace{2mm}
        \begin{subfigure}[t]{0.7\textwidth}
        \centering
        \includegraphics[width=\linewidth]{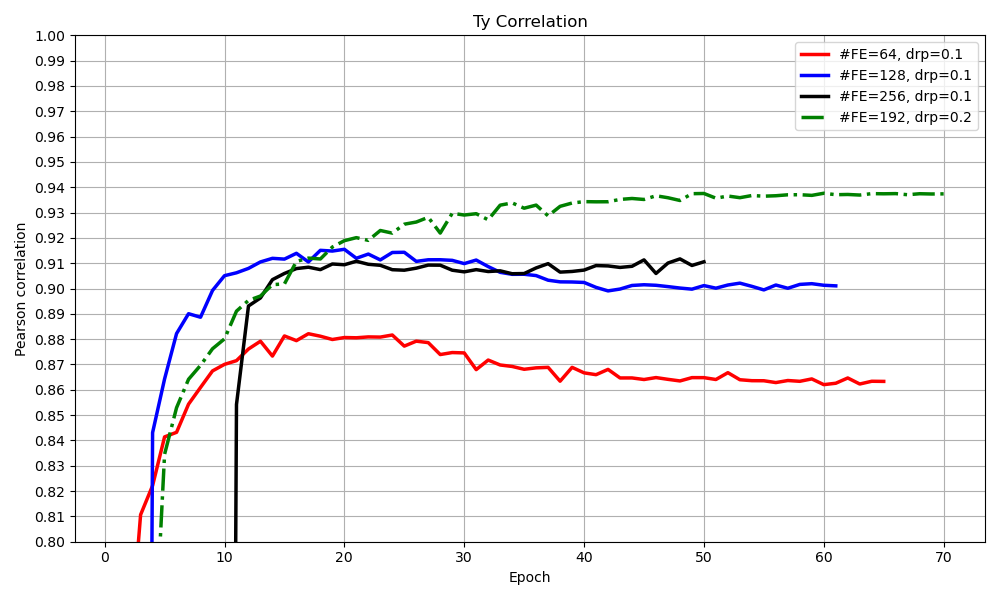}
    \end{subfigure}
    \caption{Validation-set correlation plots between predicted and ground-truth motion parameters for MoSAIK models trained with different hyperparameter configurations. The highest correlations were achieved using a 192-line crop along the frequency-encoding direction and a dropout rate of 0.2.}
    \label{plots_val_correl}
\end{figure*}

\end{document}